\documentclass[10pt,journal,compsoc]{IEEEtran}
\ifCLASSINFOpdf
\else
\fi
\usepackage[maxnames=4]{biblatex}
\usepackage{amsmath,amssymb,amsfonts}
\usepackage{algorithmic}
\usepackage{graphicx}
\usepackage{subcaption}
\usepackage{textcomp}
\usepackage{xcolor}
\usepackage{multirow}
\usepackage{tabularx,booktabs}
\usepackage{blindtext}
\usepackage[utf8]{inputenc}
\usepackage[english]{babel}
\usepackage{caption}
\usepackage{float}
\usepackage{wrapfig}
\usepackage{graphicx}

\def\BibTeX{{\rm B\kern-.05em{\sc i\kern-.025em b}\kern-.08em
    T\kern-.1667em\lower.7ex\hbox{E}\kern-.125emX}}

\begin{document}
%
\title{Exploring Facial Expressions and Affective Domains for Parkinson Detection}
%
%
%
%

\author{Luis~Felipe~Gómez-Gómez, Aythami Morales, Julian Fierrez, and Juan Rafael~Orozco-Arroyave
}

\markboth{Manuscript, December 2020}%
{Shell \MakeLowercase{\textit{et al.}}: Bare Demo of IEEEtran.cls for Computer Society Journals}
%



\IEEEtitleabstractindextext{%
\begin{abstract}

Parkinson's Disease (PD) is a neurological disorder that affects facial movements 
and non-verbal communication. Patients with PD present a reduction in facial 
movements called hypomimia which is evaluated in item 3.2 of the 
MDS-UPDRS-III scale. 
In this work, we propose to use facial expression analysis from face images based on affective domains to improve PD detection.
We propose different domain adaptation techniques to exploit the latest 
advances in face recognition and Face Action Unit (FAU) detection. The principal
contributions of this work are: 
(1) a novel framework to exploit deep face architectures to model hypomimia in PD patients; 
(2) we experimentally compare PD detection based on single images vs. image sequences while the patients are evoked various face expressions; 
(3) we explore different domain adaptation techniques to exploit existing models initially 
trained either for Face Recognition or to detect FAUs for the automatic discrimination between PD patients and healthy
subjects; and 
(4) a new approach to use triplet-loss learning to improve hypomimia modeling and PD detection. 
The results on real face images from PD patients show that we are able to properly model evoked emotions using image sequences (neutral, onset-transition, apex, offset-transition, 
and neutral) with accuracy improvements up to 5.5\% (from 72.9\% to 78.4\%) with respect to single-image PD detection. We also show that our proposed affective-domain adaptation provides  
improvements in PD detection up to 8.9\% (from 78.4\% to 87.3\% detection accuracy).

\end{abstract}

}

\maketitle

\IEEEdisplaynontitleabstractindextext

%
\IEEEpeerreviewmaketitle

\IEEEraisesectionheading{\section{Introduction}\label{sec:introduction}}

%
%
%
%
\IEEEPARstart{P}{arkinson's} Disease (PD) is a neurological disorder characterized 
by motor and non-motor impairments that affects between 1 and 2 percent of 
people over 65 years old~\cite{cacabelos2017parkinson}. 
Motor deficits include bradykinesia, rigidity, postural instability,
tremor, and dysarthria; and non-motor deficits include depression, anxiety, sleep 
disorders, and slowing of thought. 
Besides the extensive list of symptoms, most patients with PD exhibit 
also difficulties to express emotions or specific expressions on their faces.
Possible signs of those abnormalities include less range of facial muscle 
movement, wider opening of eyes, half-open mouth, and slower blinking. 
All of these phenomena in their facial expression are grouped in the 
literature and called hypomimia~\cite{bologna2013facial},
which is the result of motor impairments at the facial muscles level. 
It is typically not noticed in early stages of PD, 
but once there is a significant
deterioration, orofacial movements are highly reduced which can result in 
expressionless faces, with a very limited capability to smile, 
to express other emotions or feelings like happiness, sadness, anger, 
fear, disgust, and surprise~\cite{ekman1994strong}. 
The main effect of these impairments is in difficulties with non-verbal 
communication which also produces social isolation in a mid to long term.

Clinical evaluation of PD patients is mainly performed by expert 
neurologists according to the Movement Disorder Society - Unified 
Parkinson's Disease Rating Scale (MDS-UPDRS)~\cite{goetz2008movement}. 
This scale is the global standard for the clinical evaluation
of PD patients and it considers both motor and non-motor symptoms. 
Items of the MDS-UPDRS scale range between 0 and 4, where 0 means 
completely healthy and 4 means completely impaired.
Section III in MDS-UPDRS has a maximum value of 132 and covers motor examination including 
facial expression in one item.
According to the guidelines given by the Movement Disorder
Society, the five levels of the item where hypomimia is evaluated 
can be used to assess facial expressions in 
PD patients~\cite{goetz2008movement}. The following list indicates 
the correspondence between possible values of the item and their 
meaning in terms of facial expression evaluation:

\begin{enumerate}
    \setcounter{enumi}{-1}
    \item Normal: Normal facial expression.
    \item Slight: Minimal masked facies manifested only by decreased frequency 
    of blinking.
    \item Mild: In addition to decreased eye-blink frequency, masked facies 
    present in the lower face as well, namely fewer movements around the mouth, 
    such as less spontaneous smiling, but lips not parted.
    \item Moderate: Masked facies with lips parted some of the time when the mouth is at rest.
    \item Severe: Masked facies with lips parted most of the time when the mouth is at rest.
\end{enumerate}

Neurological evaluation highly depends on the clinician's expertise, 
which causes variability and possible bias in the rating procedure. 
Therefore, the development of computerized systems to objectively 
support the evaluation of the disease progression is now growing in importance. 
There are several contributions in the
state of the art where computerized systems are introduced to evaluate 
different aspects of Parkinson's patients including speech 
\cite{orozco2018neurospeech,moro2019phonetic}, gait~\cite{Gait,dentamaro2020gait}, handwriting~\cite{handwriting,2019_FG_PDhandw_Castrillon, 2020_COGN_HandwTrends_Faundez,de2019handwriting}, hands movement~\cite{hands}, and facial expression~\cite{bandini2017analysis}.
Among all, facial expression and hypomimia seem to be the least covered. Facial Expression Recognition (FER) refers to the evaluation of the capability
of PD patients to effectively recognize different expressions or emotions
when watching at faces. Facial Expressivity Evaluation (FEE) refers
to the capability of the patient to produce different facial expressions or
emotions. Both aspects have a very important role in social interaction and
non-verbal communication. 
The first one has been studied for several decades mainly by psychologists 
in different works and the main findings are summarized in a 
relatively recent study~\cite{Argaud2018}. 
On the other hand, FEE has become a popular field among engineers and
computer scientists, which opens space for research in different 
applications related to Affective computing. 

During the past two decades, the Affective computing community has made great advances in developing novel technologies to model facial expressions and emotional information \cite{pantic2000expert,li2018deep,2021_ICPR_Emotional_Pena}. One of the goals of affective technologies is to create computational models with the ability 
to recognize, interpret, and process human emotions, making human-computer interaction more useful. 
Sentiment analysis and affective computing have been continuously 
studied since the 20th century, helping in the development of computer vision 
systems~\cite{celiktutan2017multimodal,Li2019vision, guillen2018affective}, in the creation of entertainment~\cite{parnandi2017visual}, and in the development 
of systems to aid different areas of medicine including
neurology~\cite{wu2014objectifying,hamm2011automated,pampouchidou2017automatic}. 

Our work is focused on the study of FEE in PD patients, the main aim is to consider videos collected from patients to evaluate
their capability to produce specific emotions and to compare such a capability with respect to healthy subjects using recent advances in Affective domains.

\section{Related Works}\label{related_works}

One of the earliest studies about FEE in PD patients 
was conducted in 2004 by 
Simons et al~\cite{simons2004emotional}.
The authors evaluated the capability of 19 PD patients and 
25 healthy subjects to pose and imitate different facial expressions.
Videos with social interactions were used to evoke emotional
responses in the patients faces. The videos were manually analyzed and the participants' expressiveness was 
rated according to subjective rating scales, objective facial
measurements, and self-questionnaires. 
The objective measurement was based on the facial action coding system
presented in~\cite{ekman1978manual}, where the facial expression is decomposed according to specific facial muscle movements like rising eyebrows and wrinkling the nose. The results of the study indicated that patients with PD 
have reduced capability to produce spontaneous facial expressions in all 
experimental situations. 
Two years later in~\cite{Bowers2006}, the authors presented a work where expressivity
and bradykinesia were studied. The authors hypothesized that intentional facial
expressions are slowed (bradykinetic) and with less movement in PD patients than
in healthy controls. This hypothesis was basically inspired in other intentional
movements performed by PD patients, e.g., walking, where bradykinesia is also 
observed.
Digitized videos were evaluated frame-by-frame and the entropy in temporal 
changes of pixel intensity was measured~\cite{Richardson2000}.
The authors found that PD patients had reduced entropy compared to 
healthy controls, and were significantly slower in reaching a peak 
expression ($p<0.0001$), which is directly associated to bradykinesia.

In 2016 Almutiry et al.~\cite{Almutiry2016} presented perhaps the only 
longitudinal study about FEE in PD patients. 
A total of 8 subjects (4 PD and 4 healthy controls) participated in the study.
Patients were recorded for five days per week (once per day) during six weeks 
while controls were recorded for five days within one week.
Participants were requested to produce specific facial expressions while being recorded. The authors used two classical feature extraction methods to localise 27 facial features: 
Active Appearance Model (AAM) and Constrained Local Model (CLM).
The results suggested that PD patients exhibit less movement than 
controls, which confirms the observations made ten years earlier 
by Bowers et al.~\cite{Bowers2006}.

In 2017, Gunnery et al.~\cite{Gunnery2017} studied the coordination of movements across regions of the face in 8 PD patients (4 female). 
They used the facial action coding system~\cite{ekman1978manual,Ekman2002} to measure spontaneous facial expressions. 
The number of activated frames per action unit and their intensity was manually labeled. Correlations were computed for activation values obtained across different regions of the face. The results showed that as severity of facial expression deficit increased, there was a decrease in number, duration, intensity, and co-activation of facial muscle action.
In the same year, Bandini et al.~\cite{bandini2017analysis} classified 
emotions expressed by 17 PD patients (13 male) and equal number 
of healthy controls (6 male). 
Different emotions were evaluated including happiness, anger, disgust, 
and sadness. Different areas of the face were modeled with 49 landmarks \cite{2013PTomeFSI_FacialRegions,2018_IntelligentSystems_icb-rw},  
including: eyes, eyebrows, mouth, and nose. 
A total of 20 features were extracted to define a linear combination 
of specific reference points. Acted and imitated facial expressions were considered.
An SVM was trained to automatically detect different emotions expressed
by participants. The results with imitated expressions showed higher 
accuracies for healthy controls in most of the emotions.
The only case where the PD patients displayed an expression better than 
the healthy subjects was sadness. 
When acted expressions were evaluated, the authors found also higher accuracies
for healthy subjects than for PD patients.

Other contributions in the topic of FEE in 
PD include the study of Kang et al~\cite{kang2019voluntary}.
The authors evaluated whether deficiencies in the orofacial 
movements of PD patients occur in spontaneous and voluntary expressions.
Muscular activation (related with specific regions in the face) were
studied considering electro-myography signals.
Data from the East Asian Dynamic Facial Expression
Stimuli (EADFES) database was used~\cite{lim2013prevalence}. 
A group with 20 PD patients and 20 healthy controls was evaluated;
the authors report limitations of patients to express emotions 
spontaneously, although the observed dynamics in the movement of 
the face are similar across all subjects.
The study also highlighted the deterioration in the patient's quality of life 
due to the presence of ``masked face'', affecting social and 
psychological aspects and increasing their risk to develop
depression-related symptoms.

More recently, in another line of work, Grammatikopoulou et al.~\cite{grammatikopoulou2019detecting}
analyzed facial expressions from images captured with smartphones. 
Geometric features of the face were extracted and stored in the cloud. 
A total of 34 participants were recruited, 23 with PD and 11 healthy controls.
Patients were divided into three groups according to the facial
expression score of the MDS-UPDRS-III scale.
The authors extracted two feature sets: one by using the Google 
Face API and the other one using the Microsoft Face API \cite{2018_TIFS_SoftWildAnno_Sosa}.
The feature sets were composed by reference points on the faces,  
then two linear regression models were developed (one per feature set) 
to estimate two different values of the Hypomimia Severity index, namely HSi1 and HSi2.
These two indexes were used to classify between Parkinson's patients 
and healthy people. The reported sensitivity and specificity values were 
$0.79$ and $0.82$, respectively for HSi1 while $0.89$ and $0.73$ for HSi2.
In 2020 Sonawane and Sharma~\cite{Sonawane2020} presented a review of 
automatic techniques and the use of machine learning in detecting emotional 
facial expressions in PD patients. The authors show that the use 
of deep learning in this field has not been adequately addressed yet in the
classification between healthy people and PD patients. Also, they conducted a pilot experiment based on the use of one CNN from scratch for masked faces detection.
The pilot experiment shows that deep learning-based models can be very useful to perform the classification.

\subsection{Contributions of this Work}

As shown in the literature review, there is a lack of work in the field of 
FEE for modeling hypomimia in Parkinson's Disease (PD) patients with latest affective models including deep learning
techniques. One of the main reasons for this lack of deep approaches is the absence of large scale databases with Parkinson's Disease patients. In contrast, Face Recognition and Affective Computing research communities have made great efforts to release databases with millions of samples. In this work, we propose to use facial expression analysis and Affective domains to improve the PD detection. We propose different domain adaptation techniques \cite{2018_INFFUS_MCSreview2_Fierrez,singh2020domain} to exploit the latest developments in face recognition and Face Action Unit (FAU) detection \cite{ranjan2018deep}. The main contributions of this paper are: 
(1) a novel framework to exploit deep face architectures to model hypomimia in PD patients; 
(2) we experimentally compare PD detection based on single images vs. image sequences while the patients are evoked various face expressions; 
(3) we explore different domain adaptation techniques to exploit existing models initially 
trained either for Face Recognition or to detect FAUs for the automatic discrimination between PD patients and healthy
subjects; and 
(4) a new approach to use triplet-loss learning to improve hypomimia modeling and PD detection. 

\begin{figure*}[t]
    \centering
    \includegraphics[width=\textwidth]{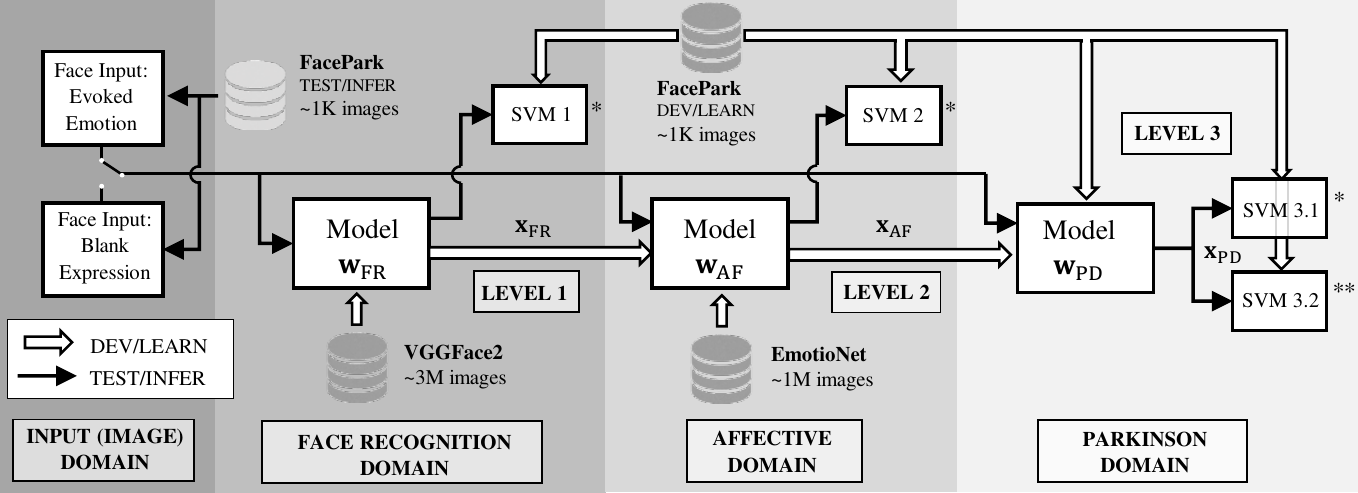}
    \caption{Experimental framework proposed for the development of this work. * SVM 1, SVM 2, and SVM 3.1 classify between PD and Healthy Control (HC); ** SVM 3.2 classify PD patients in 3 impairment levels: PD-1, PD-2, and PD-3.}
    \label{fig:Experimental-setup}
\end{figure*}

\section{Experimental Framework}


Let's assume that $\textbf{w}_{\textrm{FR}}$ is a model trained for Face Recognition tasks and the representation $\textbf{x}_{\textrm{FR}}$ is a feature vector generated by the model (typically from the last layers of a Convolutional Neural Network) from an input face image. This representation $\textbf{x}_{\textrm{FR}}$ is learned to describe the face image in a projected space where faces from the same person remain closer than faces from different persons. Similarly, models and representations can be trained for different tasks such as Affect recognition ($\textbf{w}_{\textrm{AF}}$) (e.g., in the form of facial gestures) or Parkinson's Disease detection ($\textbf{w}_{\textrm{PD}}$). Domain adaptation refers to methods that serve to adapt a representation $\textbf{x}_{\textrm{A}}$ trained for the domain $\textrm{A}$ to a new domain $\textrm{B}$ (typically a domain with similar characteristics to $\textrm{A}$ but less information to train). The resulting representation $\textbf{x}_{\textrm{B}}$, adapted from $\textbf{x}_{\textrm{A}}$,  is expected to perform better than a representation trained from scratch for the domain $\textrm{B}$.      

We propose an experimental framework where Affective features are explored at different levels (or domains). The list of domains and the corresponding underlying hypotheses to be explored are presented below. (See also Figure~\ref{fig:Experimental-setup}.)

\vspace*{3mm}

    \noindent \textbf{Face Recognition Domain (Level 1).} Our acquisition protocol introduces emotional tasks including evoked emotional states (smiling, anger, and surprise) and coordinated face gestures (right eye wink, left eye wink):
    
    \begin{itemize}
    \item \emph{Hypothesis (H1):} evoked responses intensify the features necessary to model hypomimia in Parkinson's patients. The representation $\textbf{x}_{\textrm{FR}}$ can be improved by incorporating different facial gestures during the acquisition protocol. 
    \item \emph{Experiment:} we evaluate the performance of PD detection for different sequences of face gestures using pre-trained Face Recognition models ($\textbf{w}_{\textrm{FR}}$ trained with VGGFace2 \cite{cao2018vggface2}).
    \end{itemize}
    
    \noindent \textbf{Affective Domain (Level 2).} We propose to improve
    the learned Face Recognition representations ($\textbf{x}_{\textrm{FR}}$) for Parkinson Detection by incorporating an Affective domain adaptation $\textbf{w}_{\textrm{AF}}$ training process: 
    
    \begin{itemize}
    \item \textit{Hypothesis (H2):} automatic detection of hypomimia is
    improved when features from the emotion domain are incorporated to the representations. The representation $\textbf{x}_{\textrm{AF}}$ performs better for Parkinson Detection than the representation $\textbf{x}_{\textrm{FR}}$.     
    \item \emph{Experiment:} the pre-trained models ($\textbf{w}_{\textrm{FR}}$) are adapted to the Affective domain ($\textbf{w}_{\textrm{AF}}$) using the EmotioNet database \cite{fabian2016emotionet} and FAU detection. Both, the performance of $\textbf{x}_{\textrm{FR}}$ and $\textbf{x}_{\textrm{AF}}$ are evaluated for Parkinson Detection.   
    \end{itemize}
    
    \noindent \textbf{Parkinson Domain (Level 3).} We evaluate the performance obtained by representations $\textbf{x}_{\textrm{PD}}$ trained with Healthy and Parkinson patients and the Triplet Loss function: 
    \begin{itemize}
    \item \textit{Hypothesis (H3.1):} similarity learning functions 
    designed to enhance the Parkinson features can 
    serve to improve the capability to detect hypomimia.
    \emph{Experiment:} the Affective model ($\textbf{w}_{\textrm{AF}}$) is adapted to the Parkinson domain using the Triplet Loss function and the FacePark-GITA database (see Section \ref{FacePark} for details).
    \end{itemize}
    The best  performing facial representations are also used to classify 
    patients with different levels of neurological impairment according to 
    the MDS-UPDRS-III scores:
    \begin{itemize}
    \item \textit{Hypothesis (H3.2):} facial 
    representations learned in previous models have information 
    to identify and evaluate different levels of neurological impairment 
    in Parkinson's Disease patients.
    \item \emph{Experiment:} models created to represent hypomimia are used
    to evaluate three different neurological states according to the
    MDS-UPDRS-III scores.
    \end{itemize}

\vspace*{3mm}

Details of the methods implemented to validate all hypotheses are
presented in Section~\ref{sec:methods}.

\subsection{Databases}
Three different databases are considered in this work. VGGFace2 ~\cite{cao2018vggface2} and EmotioNet ~\cite{fabian2016emotionet}
which are popular for Face Recognition and Face Action Unit detection, respectively. The third one is a new database composed by PD patients and healthy subjects. It contains face videos of patients suffering from Parkinson's
disease and age-matched healthy controls. This new corpus is called 
FacePark-GITA. Details of each database are presented below.

\subsubsection{Face Recognition Domain: VGGFace2}
This database comprises more than $3.31$ million faces from 
$9$,$131$ different subjects. An average of $362.6$ images per subject are
included~\cite{cao2018vggface2}. 
The images were downloaded from Google Image Search.
The corpus has large variations in pose, age, lighting, ethnicity, 
and profession. This database is popular in the Face Recognition community and it has been extensively used to train competitive recognition models \cite{parkhi2015face, he2016deep}.

\subsubsection{Affective Domain: EmotioNet}
This database was originally introduced by researchers from the Ohio State 
University who released the \emph{EmotioNet Challenge} 
in 2017~\cite{fabian2016emotionet}. This database 
contains one million facial expression images 
collected from the Internet. A total of $950$,$000$ images were 
annotated by the automatic Action Unit (AU) detection model presented 
in~\cite{fabian2016emotionet}, and the 
remaining $50$,$000$ images were manually annotated by experts. 
A total of $12$ AUs are included in the corpus.

\subsubsection{Parkinson Domain: FacePark-GITA} \label{FacePark}
The database was created by GITA Lab. The recording of patients is still 
ongoing and the most updated version of the corpus contains video 
recordings of 24 healthy participants and 30 PD patients. 
The videos were recorded at 30 frames per second in non-controlled 
environment conditions, i.e., light conditions and the background were not 
controlled prior the recording and differ among participants. 
PD patients were diagnosed by a neurologist expert and were evaluated 
according to the MDS-UPDRS-III scale and the Hoehn and Yahr scale (H\&Y)~\cite{goetz2004movement}. 
A summary of the clinical and demographic information is presented in 
Table~\ref{tab:Info}. 
All participants gave written informed consent. The study is 
in accordance with the Declaration of Helsinki and it
was approved by the Ethical Research Committee at the University of Antioquia.

\begin{table}[t!]
    \caption{Demographic and clinical information of the participants included in the FacePark-GITA database.}
    \centering
    \resizebox{0.48\textwidth}{!}{%
    \begin{tabular}{lcccc}
        \noalign{\hrule height 1pt}
                                    &\multicolumn{2}{c}{\textbf{PD patients}} &
                                    \multicolumn{2}{c}{\textbf{Healthy participants}} \\
                                    & Men & Women & Men & Women            \\ 
        \noalign{\hrule height 0.5pt}
           \# of Participants   & 18             & 12    & 12    & 12                 \\ 
            Age [years]         & 70.2 $\pm$ 10.4         & 67.4 $\pm$ 10.9     & 65.3 $\pm$ 8.7    & 65.2 $\pm$ 10.1  \\
            Age range [years]                      & 52 -- 90                 & 53 -- 87             & 49 -- 83           & 49 -- 80             \\ 
            $t$ [years]   & 8.7 $\pm$ 5.4         & 15.6 $\pm$ 17.3     & ---            & ---            \\ 
            $t$ range [years]    & 2 -- 20                 & 1 -- 45             & ---            & ---           \\ 
            MDS-UPDRS-III     & 35.4 $\pm$ 13.9         & 29.7 $\pm$ 12.3     & ---            & ---            \\ 
            MDS-UPDRS-III range                 & 16 -- 65                 & 15 -- 54             & ---            & ---           \\ 
            H\&Y         & 2.3 $\pm$ 0.5         & 2.5 $\pm$ 0.5     & ---            & ---            \\ 
            H\&Y range                      & 2 -- 3                   & 2 -- 3               & ---            & ---           \\ 
        \noalign{\hrule height 1pt}
        \multicolumn{5}{l}{MDS-UPDRS: Movement Disorder Society - Unified Parkinson's
        Disease Rating} \\ 
        \multicolumn{5}{l}{Scale. H\&Y: Hoehn \& Yahr scale. $t$: Years since
        diagnosis}
    \end{tabular}%
    }
    \label{tab:Info}
\end{table}

The participants of this study were asked to produce different facial
expressions while being recorded. A total of five video-task recordings
are included: right eye wink, left eye wink, smile, anger, 
and surprise. The average duration of each video is $6$ seconds.
Patients have an average age of 69 years old and healthy subjects were 
chosen with a similar range of age. 
Possible bias introduced by age or gender were discarded via 
a chi-square statistical test ($p=0.44$) and a Welch's t-test
($p=0.15$), respectively. 

\subsection{Methods}
\label{sec:methods}

\subsubsection{Face Recognition pre-trained model} \label{pretrained}

In this work we employ the ResNet50 architecture~\cite{he2016deep}, 
with 50 layers and $25.6$M parameters. This model is used to generate an initial face representation. The ResNet50 model was 
originally proposed for general image recognition tasks and 
later it was retrained with the VGGFace2 
database~\cite{cao2018vggface2} for Face Recognition. The model is used as feature extractor by removing the final decision layer. For each face image, the model generates a $1\times2048$ feature vector.

In our experiments we apply Transfer Learning (TL) \cite{pan2009survey} to adapt from one domain to another (e.g. from Face Recognition to the Affective domain). TL are methods where weights from a model originally learned
for one task are used as initialization before adjusting the model for a different task.
One of the transfer learning techniques consists in freezing
intermediate and initial layers to retain their capability to extract general 
characteristics and retrain the last layers closer to the network output. Re-training of those last layers allows to adapt the original feature space
for the new task. 
These methods are suitable for problems where data is scarce and end-to-end learning approaches fail to find the optimal feature space. The number and size of available databases to model hypomimia in patients suffering from PD are very small (typically less than $100$ subjects and less than $1$,$000$ images in total), so we expect that TL techniques will be very useful here to adapt to the Parkinson domain from the Face Recognition domain, where massive datasets are available for learning (millions of images).

\subsubsection{Face Action Unit detection models} \label{ownmodels}

In addition to the ResNet50 Face Recognition model, in this work we employ two deep neural networks trained from scratch for Face Action Unit (FAU) detection. The architectures employed are based on the popular VGG and ResNet models \cite{yan2019vargfacenet,ranjan2018deep}. The details of the two models are described below:    

\textit{VGG-8}: This model contains 8 convolutional layers divided 
into groups of 2 layers. Each group is followed by a Max pooling layer.
Convolutional layers apply a variety of filters to the images and 
Max-Pooling layers reduce the size of the filtered images. 
Additionally, dropout is used in the regularization layers
to randomly discard neurons in the model and make it less prone to
overfitting. The final part of the architecture has a total of 
six convolutional layers (fully-connected) before the decision 
layer. The number of neurons per layer is $1024$, $512$, $256$, $128$, $64$, and $32$. The number of parameters of this model is $295$,$448$.

\textit{ResNet-7}: The ResNet model is composed of a total of 7 residual blocks. Each block can be defined as an identity-block or a conv-block. The identity-blocks are the standard blocks used in ResNet, they have a set of convolutional filters and a shortcut connection which bypasses these blocks. This block has the same input and output dimensions. Conv-blocks are the block types where the input and output dimensions do not match. The difference with the identity-block is a convolutional layer in the shortcut to the output. The benefit of these architectures is that in traditional architectures by having 
a high amount of layers in the training, the problem of error degradation appears. ResNet models with their previous layer shortcut connections are effective in solving this problem~\cite{he2016deep}. The number of parameters of this model is $366$,$626$.

\subsubsection{Affective Triplet Loss}

Due to the limited number of samples in the FacePark-GITA database, for the Parkinson domain adaptation we opted for a Triplet Loss learning approach. The Triplet Loss function  consists in applying a linear transformation
over the data before taking the distance among samples.
Given a training data set $\mathcal{S} = \left (\textbf{x}_i, y_i \right)$ 
with inputs $\textbf{x}_i \in \mathbb{R}^d$ and discrete class 
labels $y_i \in \mathbb{Z}$, the goal is to find a transformation 
to the input data such that reduces the distance between pairs 
from the same class while increases the distance between pairs from 
different classes. 
The Mahalanobis distance defined in Equation~\ref{eq:KNN_M} is the 
similarity measure used in this work. 

\begin{equation} \label{eq:KNN_M}  
     d^2_\mathbf{M}(\textbf{x}_i,\textbf{x}_j) = (\textbf{x}_i - \textbf{x}_j)^T \mathbf{M} (\textbf{x}_i - \textbf{x}_j)
\end{equation}

\noindent where $\mathbf{M}$ is a positive semi-defined symmetric matrix that 
can be decomposed as $\mathbf{M}=\mathbf{T}^T \mathbf{T}$, where 
$\mathbf{T}$ denotes a linear transformation matrix. 
Equation~\ref{eq:KNN_M} can be rewritten as:

\label{eq:KNN_Lineal}  
\begin{alignat}{3}
    d^2_\mathbf{M}(\textbf{x}_i,\textbf{x}_j) & = (\mathbf{T}(\textbf{x}_i - \textbf{x}_j))^T \mathbf{T} \left(\textbf{x}_i - \textbf{x}_j \right ) \\
    & = \left \| \mathbf{T}\left(\textbf{x}_i - \textbf{x}_j \right ) \right \|_2^2 = \left \| {\textbf{x}_i}' - {\textbf{x}_j}' \right \|_2^2
\end{alignat}

The linear transformation $\mathbf{T}$ can be generalized as
$\Phi(\textbf{x}_i)$, where $\Phi$ indicates 
a kernel function. The resulting distance metric is as follows:
 
\begin{equation} \label{eq:KNN_No_Lineal}  
     d^2_\mathbf{M}(\textbf{x}_i,\textbf{x}_j) = \left \| {\Phi(\textbf{x}_i)} - {\Phi(\textbf{x}_j)} \right \|_2^2
\end{equation}

The process to determine the transformed vector $\Phi(\textbf{x})$, 
requires to find a transformation that makes the 
intra-class distance smaller than the inter-class distance. 
The general rule which is applied over the data set consists
in the following triplets $\mathcal{S}_{\mathrm{T}}$: 

\begin{equation}
   \mathcal{S}_{\mathrm{T}} = \{(\textbf{x}^a,y^a),(\textbf{x}^n,y^n),(\textbf{x}^p,y^p) | y^a = y^p, y^a \neq  y^n\}
\end{equation}

\noindent where ${a,p}$ are samples belonging to the same class, and ${n}$ is a sample from a different class. In our Parkinson detection experiments, the number of classes is two (healthy and Parkinson). However, we propose to introduce an additional restriction in the triplet. In our experiments, ${a,p}$ belong to the same class, but present different face expression. in this way, we introduce facial gestures into the learning objective. The generation of the triplet $\mathcal{S}_{\mathrm{T}}$ can be seen as a data augmentation technique. The high number of possible combinations of three elements in a dataset enriches the training process, especially when low number of samples are available.  The triple loss function to be minimized is defined as:
\begin{equation} \label{eq:triplet}  
     \mathcal{L} =  \sum_{\mathcal{S}_{\mathrm{T}}}
     [d^2_\mathbf{M}(\textbf{x}^a,\textbf{x}^p) - d^2_\mathbf{M}(\textbf{x}^a,\textbf{x}^n) + \alpha ]_+
\end{equation}

\noindent where $[z]_+ = \max(z,0)$, and $\alpha \geq 0$ is the minimum margin required 
between classes.

\subsection{Classification and Parameter Optimization}
\label{sec:classification}

The automatic classification between healthy people and PD patients is performed using 
Support Vector Machines (SVMs). The classification of patients with different degree
of impairment is performed using SVMs optimized in a one vs. all strategy.
In the binary classification experiments with SVMs, linear and Gaussian 
kernels are considered.
The optimization of hyper-parameters is performed in a search grid up to
powers of ten with $C \in \{10^{-4}, 10^{-3}, 10^{-2}, 
\ldots, 10^{2}, 10^{3} \}$ and 
$ \gamma \in \{ 10^{-4}, 10^{-3}, 10^{-2}, \ldots , 10^{3} \}$ 
for the Gaussian kernel, and for the linear kernel the search considered 
$C \in \{  10^{-1}, 1, 10^{1}, 10^{2}, 10^{3}, 10^{4}$\}. 
In the multi-class classification only linear kernels were considered.
Optimization and evaluation of the models is performed following a 
5-folds cross-validation strategy. Results of the binary classification
are reported in terms of accuracy (Acc), sensitivity (Sens), 
specificity (Spec), F1-Score (F1), and Area Under the receiver 
operating characteristic Curve (AUC). Results of the multi-class classification
are reported in terms of accuracy (Acc), F1-Score (F1), Kappa coefficient 
($\kappa$), and confusion matrix.
In all of the cases, results include values of the optimal 
hyper-parameters which are found as the mode along the parameters 
considered along the test folds of each experiment.

\section{Experiments and Results}

\begin{figure*}[ht]
    \centering
    \includegraphics[width=\textwidth]{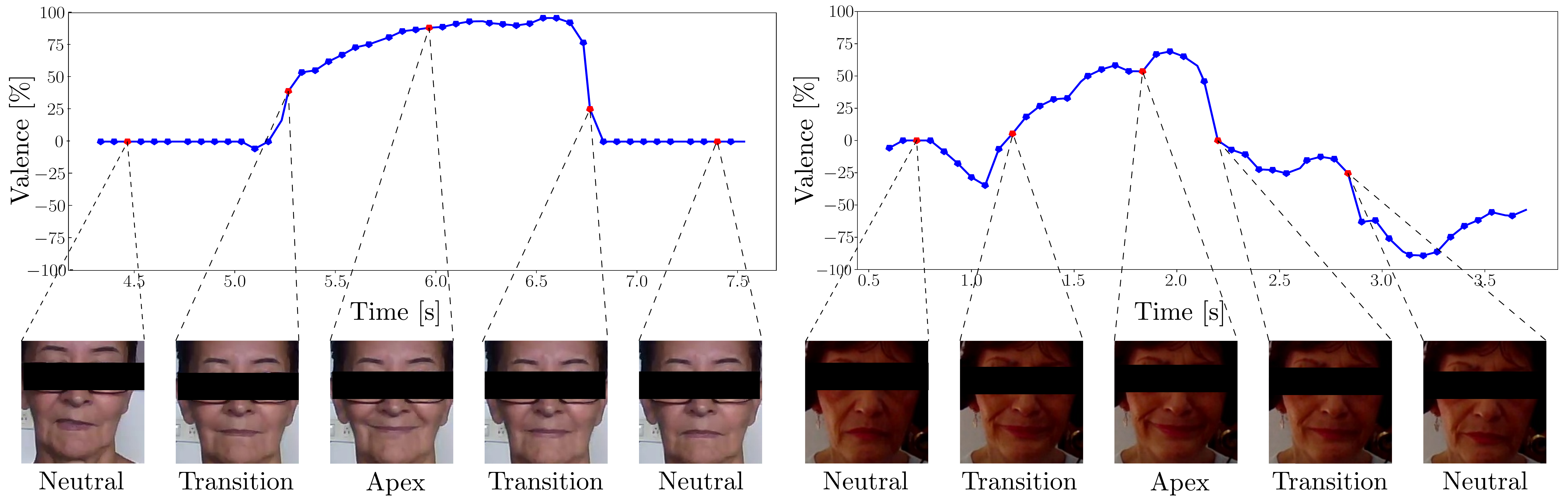}
    \caption{Emotion stages according to the evoked valence measured with the Affectiva tool. (left) Healthy woman 63 years old; (right) Woman with Parkinson's disease, 67 years old, facial expression item = 2.}
    \label{fig:stages_emotions}
\end{figure*}

FacePark-GITA includes 5 videos for each participant. Each video corresponds to a different facial expression: 
smile, anger, surprise, left eye wink, and right eye wink. 
Five frames per video-task were extracted with the
software Affectiva\footnote{https://www.affectiva.com/}. 
The curve of valence provided by the software is used as the criterion to
select the following sequence of five images/frames per participant on each expression:
(i) Neutral; (ii) transition from Neutral to the Apex (i.e., onset);
(iii) Apex; (iv) transition from the Apex to Neutral (i.e., offset); and (v) Neutral.
The sequence of images and their direct relation with the valence curve are 
illustrated in Figure~\ref{fig:stages_emotions}.

\subsection{Experiment 1: Face Recognition Domain}

\subsubsection{PD detection based on single face images}

Individual frames corresponding to each valence level shown in
Figure~\ref{fig:stages_emotions}
are considered to evaluate whether specific frames provide relevant information to
discriminate between PD patients and healthy subjects.
Feature vectors are obtained from the last layer of the ResNet50 model (see Section \ref{pretrained}). 
Table~\ref{tab:OneFrameVGGFace} summarizes the results.

\begin{table}[ht]
    \caption{Results of classification using a single image from the extracted image sequence.}
    \centering
    \resizebox{.48\textwidth}{!}
    {%
    \begin{tabular}{llllll}
        \noalign{\hrule height 1pt}
        \textbf{E.S.} & \textbf{Kernel*} & \textbf{Acc[\%]}    & \textbf{Sens[\%]} & \textbf{Spec[\%]}  & \textbf{F1[\%] }      \\
        \noalign{\hrule height 0.5pt}
        Neutral  & $C$=1e+01; $\gamma$=1e-04 	 & 	69.0 $\pm$ 10.1 	 & 	 74.0 $\pm$ 11.6 	 & 	 63.0 $\pm$ 9.7 	 & 	 67.8 $\pm$ 10.1 \\
        Apex     & $C$=1e+01; $\gamma$=1e-04 	 & 	70.0 $\pm$ 9.1 	 & 	 84.4 $\pm$ 7.9 	 & 	 53.3 $\pm$ 24.0 	 & 	 61.0 $\pm$ 18.6 \\
        Onset    & $C$=1e+01; $\gamma$=1e-04 	 & 	71.4 $\pm$ 3.2 	 & 	 88.6 $\pm$ 7.0 	 & 	 50.0 $\pm$ 9.0 	 & 	 63.1 $\pm$ 6.6 \\
        Offset   & $C$=1e+01; $\gamma$=1e-04 	 & 	71.6 $\pm$ 5.2 	 & 	 79.5 $\pm$ 3.3 	 & 	 61.9 $\pm$ 13.5 	 & 	 68.6 $\pm$ 8.2 \\
        \noalign{\hrule height 0.5pt}
        Neutral  & $C$=1e-03 	 & 	70.8 $\pm$ 9.6 	 & 	 77.3 $\pm$ 10.2 	 & 	 63.0 $\pm$ 9.7 	 & 	 69.3 $\pm$ 9.7 \\
        Apex     & $C$=1e-03 	 & 	70.8 $\pm$ 9.1 	 & 	 83.7 $\pm$ 7.3 	 & 	 55.7 $\pm$ 21.6 	 & 	 63.8 $\pm$ 16.3 \\
        \textbf{Onset}    & \textbf{$C$=1e-02} 	 & 	\textbf{72.9 $\pm$ 4.2} 	 & 	 \textbf{88.6 $\pm$ 7.8} 	 & 	 \textbf{53.4 $\pm$ 7.7} 	 & 	 \textbf{66.1 $\pm$ 5.9} \\
        Offset   & $C$=1e-01 	 & 	72.8 $\pm$ 4.3 	 & 	 81.5 $\pm$ 4.5 	 & 	 61.9 $\pm$ 13.5 	 & 	 69.2 $\pm$ 7.9 \\
        \noalign{\hrule height 1pt}
        \multicolumn{6}{l}{\textbf{E.S.}: Expression stage. First three rows: Gaussian kernel. Last three rows: Linear kernel.}\\
        \multicolumn{6}{l}{*Column with optimal hyper-parameters.}
        \label{tab:OneFrameVGGFace}
    \end{tabular}%
    }
\end{table}
Note that there is almost no difference among the accuracies obtained with the
frames of each expression stage. Perhaps the only thing to highlight is the high
sensitivity (88,6\%) of the Onset stage, which likely indicates that this
stage is maybe a good choice to model hypomimia in specific frames within a video. 
This preliminary observation will be further elaborated in the next 
experiments.
%
%

\subsubsection{PD detection based on image sequences}

Given the small amount of information provided by individual frames, 
we evaluate the use of multi-frame sequences in a simple information fusion architecture based score fusion \cite{2018_INFFUS_MCSreview1_Fierrez} as a way to capture changes during the production of facial expressions. The general idea was already studied in~\cite{Orozco2016} for speech signals, where the author hypothesised
that PD patients have more difficulties to start or stop the movement of
muscles and limbs during speech production. The idea was later
extended to other movements like handwriting and gait~\cite{Vasquez2019}.

As in the case of speech, gait, and handwriting, we believe that the same 
hypothesis holds during the production of facial expressions. 
Thus, the analysis of multiple-frames during the production of facial expressions should provide useful information to discriminate between PD patients and healthy subjects. The following multi-frame sequences are considered:

\begin{itemize}
    \item NOnA: Neutral, Onset, and Apex.
    \item AOffN: Apex, Offset, and Neutral.
    \item NOnAOffN: Neutral, Onset, Apex, Offset, and Neutral.
\end{itemize}

Table~\ref{tab:SequenceFramesVGGFace} shows the results obtained when 
the changes in the production of facial expressions are incorporated by
feature vectors extracted from multi-frame sequences.

\begin{table}[t!]
    \caption{Results of the classification using different combinations of the extracted frames sequences}
    \centering
    \resizebox{.48\textwidth}{!}
    {
    \begin{tabular}{llllll}
        \noalign{\hrule height 1pt}
        \textbf{Sequences} & \textbf{Kernel*} & \textbf{Acc[\%] }   & \textbf{Sens[\%]} & \textbf{Spec[\%]}  & \textbf{F1[\%]}      \\
        \noalign{\hrule height 0.5pt}
        NOnA & $C$=1e+02; $\gamma$=1e-04 	 & 	77.4 $\pm$ 8.7 	 & 	 89.3 $\pm$ 4.6 	 & 	 63.0 $\pm$ 16.1 	 & 	 72.9 $\pm$ 11.2 \\
        AOffN & $C$=1e+01; $\gamma$=1e-04 	 & 	76.3 $\pm$ 8.0 	 & 	 86.8 $\pm$ 12.0 	 & 	 63.5 $\pm$ 22.4 	 & 	 69.2 $\pm$ 17.8 \\
        NOnAOffN & $C$=1e+01; $\gamma$=1e-04 	 & 	77.2 $\pm$ 8.6 	 & 	 86.1 $\pm$ 14.8 	 & 	 67.2 $\pm$ 10.3 	 & 	 74.2 $\pm$ 8.5 \\
        \noalign{\hrule height 0.5pt}
        NOnA & $C$=1e-03 	 & 	78.2 $\pm$ 9.8 	 & 	 90.1 $\pm$ 5.2 	 & 	 63.8 $\pm$ 17.1 	 & 	 73.8 $\pm$ 12.6 \\
        AOffN & $C$=1e-03 	 & 	77.8 $\pm$ 9.1 	 & 	 88.8 $\pm$ 9.4 	 & 	 64.2 $\pm$ 24.1 	 & 	 70.4 $\pm$ 20.5 \\
        \textbf{NOnAOffN} & \textbf{$C$=1e-03} 	 & 	\textbf{78.4 $\pm$ 7.1} 	 & 	 \textbf{87.8 $\pm$ 11.4} 	 & 	 \textbf{67.7 $\pm$ 11.6} 	 & 	 \textbf{75.4 $\pm$ 7.9} \\
        \noalign{\hrule height 1pt}
        \multicolumn{6}{l}{First three rows: Gaussian kernel. Last three rows: Linear kernel.}\\
        \multicolumn{6}{l}{*Column with optimal hyper-parameters.}
        \label{tab:SequenceFramesVGGFace}
    \end{tabular}%
    }
\end{table}

The results obtained by the  affective sequences are better than those obtained with individual
frames. The improvement is around 7\% and the best result is obtained with the two
cases where the sequence NOnA is included, which is focused on 
modeling information in the transition
between neutral and the production of a certain expression.
It is also worth to highlight that sensitivity is near 90\% in all of the 
cases, while specificity is rather low (around 64\%). 
This indicates that the proposed approach is good to detect patients but 
not as good to detect healthy controls.

This result validates the hypothesis \textit{H1} about the existence of useful information related to hypomimia
in the evoked facial expressions.
Given this clear improvement, the next experiments will include only
feature vectors extracted from multi-frame sequences.

\subsection{Experiment 2: Affective Domain}

\begin{figure}[t]
\captionsetup[subfigure]{labelformat=empty, justification=centering}
\centering

    \begin{subfigure}{0.2\textwidth}
      \centering
      \includegraphics[width=.6\linewidth]{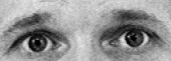}  
      \caption{AU1: Inner Brown \\ Raiser}
    \end{subfigure}
    \begin{subfigure}{0.2\textwidth}
      \centering
      \includegraphics[width=.6\linewidth]{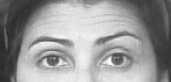}  
      \caption{AU2: Outer Brown \\ Raiser}
    \end{subfigure}
    
    \begin{subfigure}{0.2\textwidth}
      \centering
      \includegraphics[width=.6\linewidth]{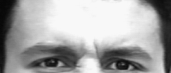}  
      \caption{AU4: Brow Lowerer}
    \end{subfigure}
    \begin{subfigure}{0.2\textwidth}
      \centering
      \includegraphics[width=.6\linewidth]{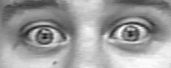}  
      \caption{AU5: Upper Lid \\ Raiser}
    \end{subfigure}
    
    \begin{subfigure}{0.2\textwidth}
      \centering
      \includegraphics[width=.6\linewidth]{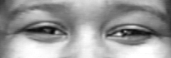}  
      \caption{AU6: Check Raiser}
    \end{subfigure}
    \begin{subfigure}{0.2\textwidth}
      \centering
      \includegraphics[width=.6\linewidth]{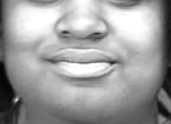}  
      \caption{AU12: Lip Corner \\ Puller}
    \end{subfigure}

    \begin{subfigure}{0.2\textwidth}
      \centering
      \includegraphics[width=.6\linewidth]{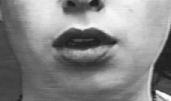}  
      \caption{AU25: Lips Part}
    \end{subfigure}
    \begin{subfigure}{0.2\textwidth}
      \centering
      \includegraphics[width=.6\linewidth]{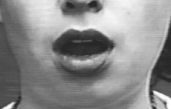}  
      \caption{AU26: Jaw Drop}
    \end{subfigure}

    \caption{Action Units defined for the Experiment 2. Source: \cite{ekman1978manual}.}
    \label{fig:AU}
\end{figure}

This experiment intends to incorporate information from the Affective 
domain to improve Parkinson's Disease (PD) detection. 
In this case the EmotioNet database is used to create an appropriate facial 
representation space. The first step consists in selecting AUs that provide suitable information to perform the automatic classification between PD patients and healthy subjects. We selected a subset of AUs according to~\cite{Ekman2002} adequate for the facial expressions included in the recording tasks of the FacePark-GITA database. Figure~\ref{fig:AU} shows the set of selected AUs.

\begin{table*}[t]
\caption{FAU detection results of the VGGFace2 model after retraining with the EmotioNet database.}
\centering
\begin{tabular}{llcccccccc}
    \noalign{\hrule height 1pt}
        
        \textbf{Models} & \textbf{Metrics} & \textbf{AU 1}  & \textbf{AU 2}  & \textbf{AU 4}  & \textbf{AU 5}  & \textbf{AU 6}  & \textbf{AU 12}  & \textbf{AU 25}  & \textbf{AU 26}  \\
         \noalign{\hrule height 0.5pt}
        \multirow{2}{*}{Baseline ($\textbf{x}_{\textrm{FR}}$)} & AUC & 0.83 & 0.83 & 0.87 & 0.80 & 0.94 & 0.95 & 0.92 & 0.80 \\
        & EER [\%] & 24.58 & 23.78 & 21.01 & 27.13 & 12.82 & 12.11 & 15.38 & 27.32 \\
    \noalign{\hrule height 0.5pt}
        \multirow{2}{*}{Freeze 75 ($\textbf{x}_{\textrm{AF}}$)}  & AUC  & 0.84 & 0.84 & 0.86 & 0.84 & 0.92 & 0.93 & 0.95 & 0.85 \\
        & EER [\%] & 21.84 & 20.80 & 19.90 & 21.65 & 14.34 & 10.42 & 8.63 & 22.48 \\
    \noalign{\hrule height 0.5pt}
        \multirow{2}{*}{Freeze 50 ($\textbf{x}_{\textrm{AF}}$)} & AUC  & 0.84 & 0.87 & 0.87 & 0.87 & 0.93 & 0.95 & 0.90 & 0.83 \\
        & EER [\%] & 20.56 & 19.29 & 18.92 & 19.53 & 13.22 & 10.58 & 10.99 & 24.32 \\
    \noalign{\hrule height 1pt}
\end{tabular}%
\label{tab:VGG_Emotionet}
\end{table*}

\subsubsection{Adaptation from Face Recognition models}

The process to adapt the convolutional models from one domain to another consists 
in freezing different percentages of the layers and retraining the remaining
portion. The data with the selected AUs from the EmotioNet dataset are used
here to retrain the models.
In this case we evaluate three percentages of layers frozen during the retraining of the ResNet50 model (originally trained for Face Recognition): freezing 50\% (Freeze 50), 
freezing 75\% (Freeze 75), and freezing 100\%. 
Note that the freezing 100\% model is taken as the Baseline and corresponds to the case where no affective information is 
incorporated ($\textbf{x}_{\textrm{FR}}$).
After the convolutional layers, a fully connected layer is added for the 
classification of the 8 selected AUs (see Figure~\ref{fig:AU}). The result of the retraining process and its performance to classify the AUs is shown in Table~\ref{tab:VGG_Emotionet} in 
terms of AUC and EER values. The accuracy varies depending of the FAU and the 
percentage of layers frozen. The FAUs numbers $6$, $12$, and $25$ reached accuracies around $90\%$, while the rest of the FAUs achieved performances around $80\%$.

The representations $\textbf{x}_{\textrm{AF}}$ obtained by the retrained models are further used to classify between PD patients and healthy 
subjects of the FacePark-GITA corpus.
The results obtained with the Freeze 75 and Freeze 50 models are shown in 
Table~\ref{tab:Emotion75} and Table~\ref{tab:Emotion50}, respectively. 
The results for the Baseline model correspond to those previously shown in Table~\ref{tab:SequenceFramesVGGFace}. Optimal hyper-parameters found in the
5-fold cross-validation process are also included in every experiment.

\begin{table}[t]
    \caption{PD classification results using the Freeze 75 model.}
    \centering
    \resizebox{.48\textwidth}{!}
    {%
    \begin{tabular}{llllll}
    \noalign{\hrule height 1pt}
    \textbf{Sequence} & \textbf{Kernel*} & \textbf{Acc[\%] }   & \textbf{Sens[\%]} & \textbf{Spec[\%]}  & \textbf{F1[\%]}      \\
        \noalign{\hrule height 0.5pt}
        NOnA & $C$=1e+01; $\gamma$=1e-04 	 & 	84.2 $\pm$ 5.4 	 & 	 90.0 $\pm$ 8.3 	 & 	 77.2 $\pm$ 10.8 	 & 	 82.3 $\pm$ 6.3\\
        AOffN & $C$=1e+02; $\gamma$=1e-04 	 & 	81.6 $\pm$ 8.6 	 & 	 87.8 $\pm$ 7.4 	 & 	 73.9 $\pm$ 11.5 	 & 	 80.0 $\pm$ 9.5 \\
        NOnAOffN & $C$=1e+02; $\gamma$=1e-04 	 & 	86.7 $\pm$ 8.9 	 & 	 91.2 $\pm$ 4.7 	 & 	 81.6 $\pm$ 15.5 	 & 	 85.5 $\pm$ 10.2 \\
        \noalign{\hrule height 0.5pt}
        NOnA & $C$=1e-01 	 & 	84.7 $\pm$ 5.4 	 & 	 89.5 $\pm$ 9.4 	 & 	 78.9 $\pm$ 11.3 	 & 	 82.9 $\pm$ 6.5 \\
        AOffN & $C$=1e-01 	 & 	82.6 $\pm$ 9.6 	 & 	 87.8 $\pm$ 8.3 	 & 	 76.1 $\pm$ 13.3 	 & 	 81.2 $\pm$ 10.4 \\
        \textbf{NOnAOffN} & \textbf{$C$=1e-01} 	 & 	\textbf{87.3 $\pm$ 8.0} 	 & 	 \textbf{90.6 $\pm$ 5.0 }	 & 	 \textbf{83.6 $\pm$ 13.1} 	 & 	 \textbf{86.6 $\pm$ 8.8} \\
        \noalign{\hrule height 1pt}
        \multicolumn{6}{l}{First three rows: Gaussian kernel. Last three rows: Linear kernel.}\\
        \multicolumn{6}{l}{*Column with optimal hyper-parameters.}
    \end{tabular}%
    }
    \label{tab:Emotion75}
\end{table}

\begin{table}[t]
    \caption{PD classification results using the Freeze 50 model.}
    \centering
    \resizebox{.48\textwidth}{!}{%
    \begin{tabular}{llllll}
        \noalign{\hrule height 1pt}
        \textbf{Sequence} & \textbf{Kernel*} & \textbf{Acc[\%] }   & \textbf{Sens[\%]} & \textbf{Spec[\%]}  & \textbf{F1[\%]}      \\
        \noalign{\hrule height 0.5pt}
        \textbf{NOnA} & \textbf{$C$=1e+01; $\gamma$=1e-04}	 & 	\textbf{83.1 $\pm$ 6.0}	 & 	 \textbf{87.7 $\pm$ 12.4} 	 & 	 \textbf{77.5 $\pm$ 10.2} 	 & 	 \textbf{81.1 $\pm$ 6.5} \\
        AOffN & $C$=1e+01; $\gamma$=1e-04 	 & 	81.3 $\pm$ 7.5 	 & 	 86.3 $\pm$ 13.0 	 & 	 75.6 $\pm$ 3.6 	 & 	 80.1 $\pm$ 6.8 \\
        NOnAOffN & $C$=1e+00; $\gamma$=1e-04 	 & 	81.9 $\pm$ 9.2 	 & 	 97.4 $\pm$ 2.5 	 & 	 63.4 $\pm$ 17.7 	 & 	 75.5 $\pm$ 14.3 \\
        \noalign{\hrule height 0.5pt}
        NOnA & $C$=1e-01 	 & 	82.1 $\pm$ 6.8 	 & 	 85.0 $\pm$ 13.8 	 & 	 78.6 $\pm$ 11.0 	 & 	 80.2 $\pm$ 7.7 \\
        AOffN & $C$=1e-01 	 & 	80.0 $\pm$ 7.6 	 & 	 83.4 $\pm$ 12.7 	 & 	 76.1 $\pm$ 4.4 	 & 	 79.1 $\pm$ 7.2 \\
        NOnAOffN & $C$=1e-01 	 & 	80.2 $\pm$ 11.1 	 & 	 84.3 $\pm$ 8.5 	 & 	 75.3 $\pm$ 19.1 	 & 	 78.3 $\pm$ 13.0\\
        \noalign{\hrule height 1pt}
        \multicolumn{6}{l}{First three rows: Gaussian kernel. Last three rows: Linear kernel.}\\
        \multicolumn{6}{l}{*Column with optimal hyper-parameters.}
    \end{tabular}%
    }
    \label{tab:Emotion50}
\end{table}

\begin{figure*}[t]
\centering

    \begin{subfigure}{0.3\textwidth}
      \centering
      \includegraphics[width=1\linewidth]{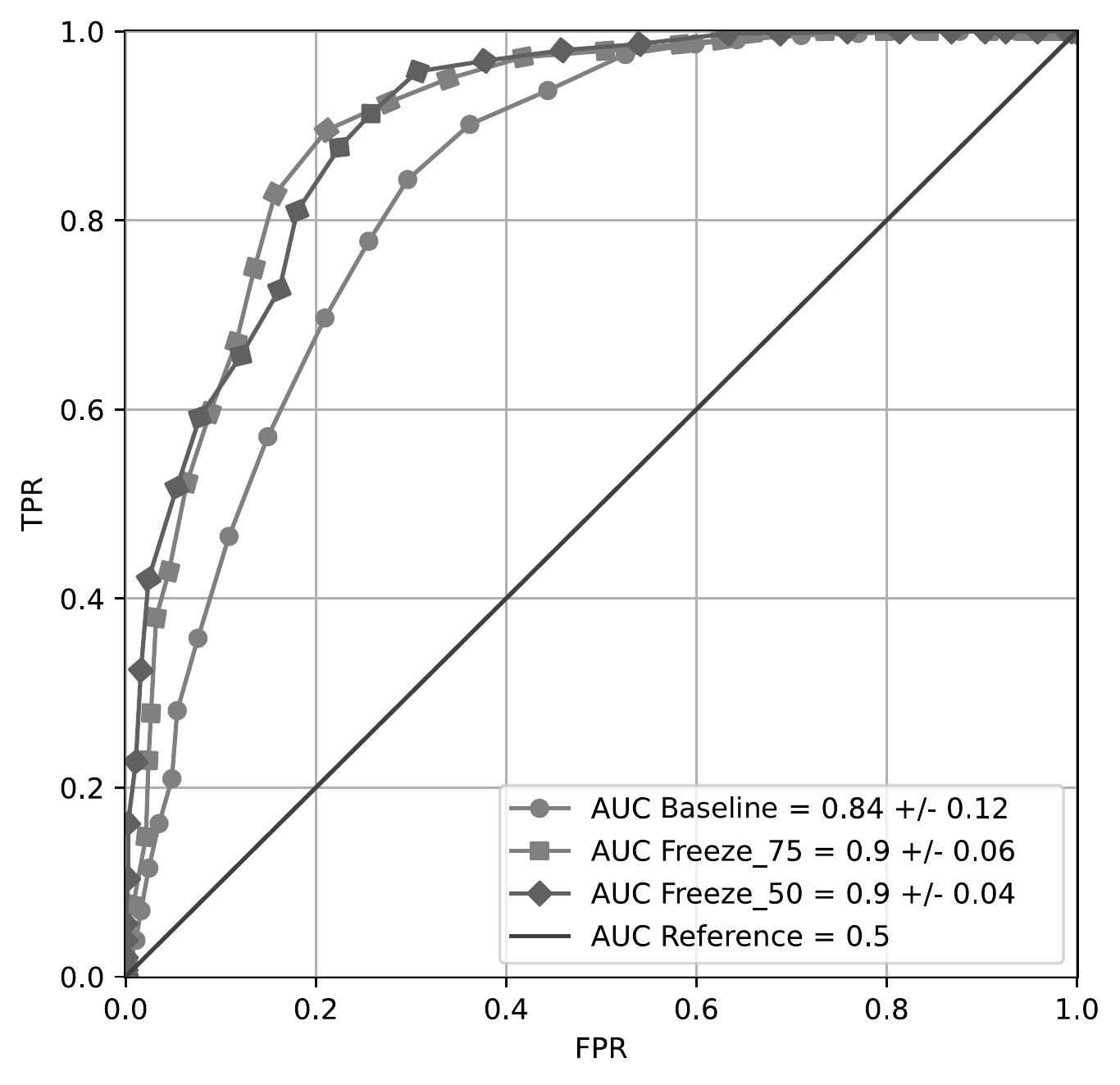}  
      \caption{NOnA.}
    \end{subfigure}
    \begin{subfigure}{0.3\textwidth}
      \centering
      \includegraphics[width=1\linewidth]{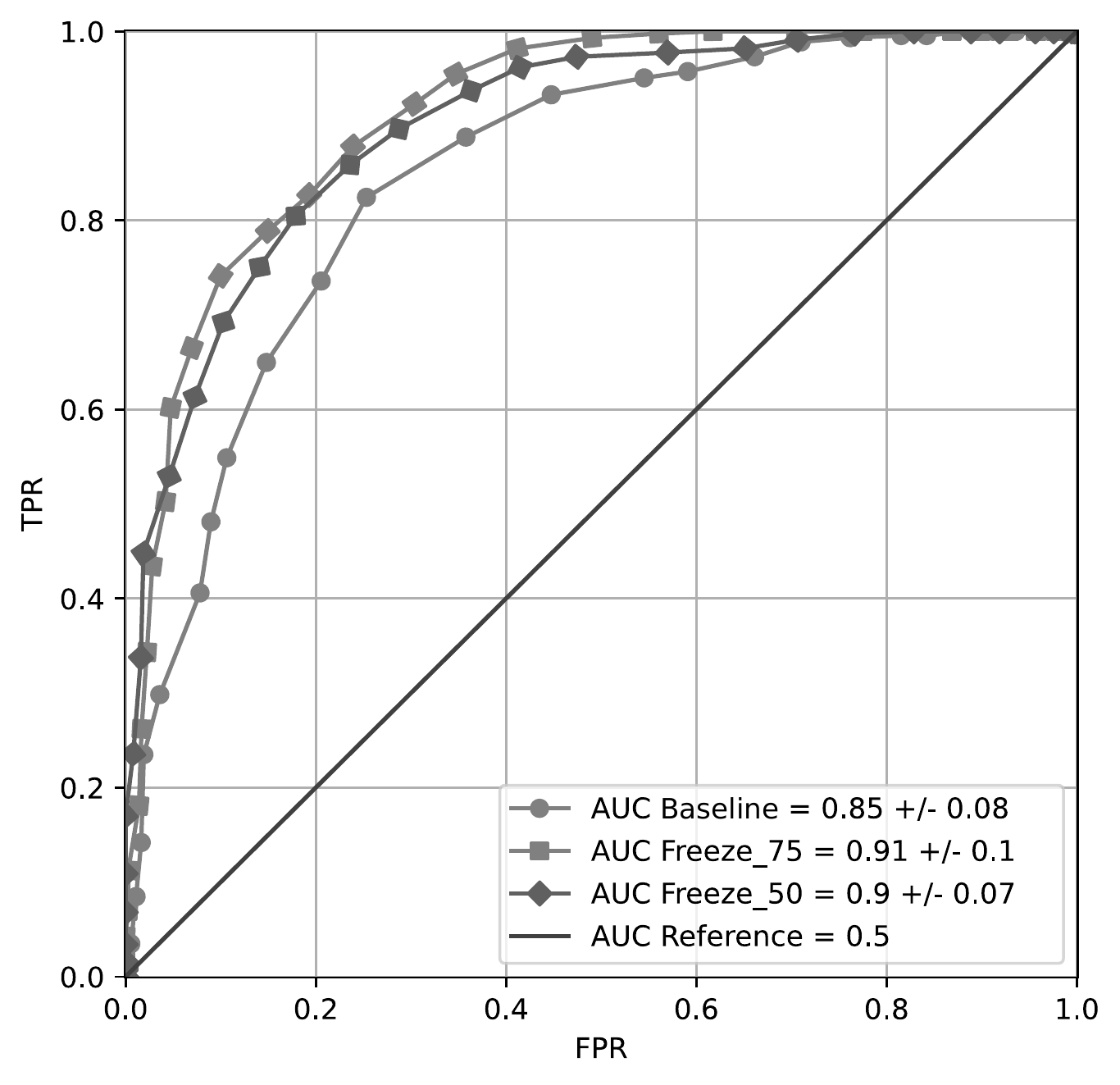}  
      \caption{AOffN.}
    \end{subfigure}
    \begin{subfigure}{0.3\textwidth}
      \centering
      \includegraphics[width=1\linewidth]{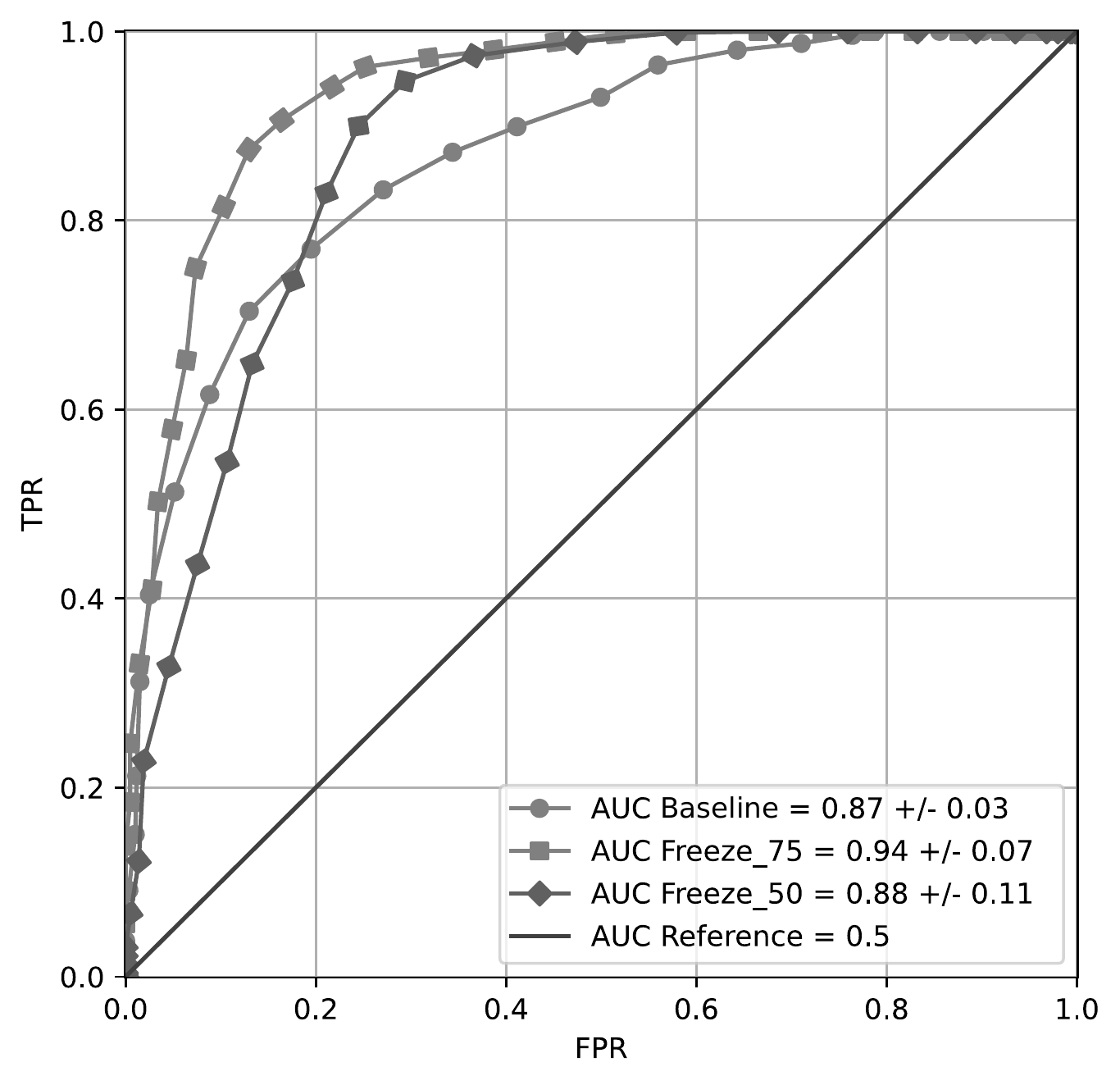}  
      \caption{NOnAOffN.}
    \end{subfigure}

    \caption{PD classification ROC curves obtained from the different input sequences in the retrained Freeze models.}
    \label{fig:EmotionSC}
\end{figure*}

Note that the Freeze 75 exhibits higher accuracies than the Freeze 50, indicating 
that considerable information from the Face Recognition domain is still useful
to obtain good results in the classification between PD patients and 
healthy subjects. 
More interestingly, note that the best accuracy obtained with the Freeze 75 model
in Table~\ref{tab:Emotion75} ($87.3\%$) is $8.9\%$ higher than the best result 
obtained when only a Face Recognition model is considered (Table~\ref{tab:SequenceFramesVGGFace}). 
This result supports our second hypothesis (\textit{H2}), the idea of
incorporating information from the Affective domain to the Face Recognition domain
 to improve detection of hypomimia in PD patients.
The benefits of including information of the 
Affective domain are also shown in Figure~\ref{fig:EmotionSC}, where the 
ROC curves obtained with the Freeze 75, Freeze 50, and Baseline models
are presented.

Note that the models used until this point of the study are based on 
architectures originally trained for Face Recognition tasks (ResNet50). 
Now we want to evaluate the importance of this initialization based on a Face Recognition training processes.

\subsubsection{Training Affective models from scratch}

The previous scenario studied the performance of pre-trained models with high number of parameters learned from the Face Recognition domain after adaptation to the Affective domain. In this section we will train FAU detection models from scratch. ResNet50 requires to optimize more than $20$M parameters. 
Conversely, the VGG-8 and ResNet-7 architectures proposed in Section \ref{ownmodels} require the optimization of 
$295$,$448$ and $366$,$626$ parameters respectively. 
These reduced architectures are trained with the same data as those
considered previously to retrain the Freeze 50 and Freeze 75 models.
Table~\ref{tab:CNN-Tiny} shows the results with the AUC values obtained 
when the different AUs are detected.
Note that these results are higher than those reported 
in Table~\ref{tab:VGG_Emotionet} where greater number of parameters
are optimized. However, the ResNet50 was originally trained for Face Recognition tasks, where face gestures
are features to be excluded from the representation space.
This result indicates that a simpler model might provide high enough 
AUs discrimination performance to be used in the classification
between PD patients and healthy controls.

\begin{table}[t]
\caption{FAU detection results of the VGG-8 and ResNet-7 training with EmotioNet database}
\centering
\resizebox{\columnwidth}{!}
    {%
    \begin{tabular}{llcccccccc}
        \noalign{\hrule height 1pt}
            \textbf{Models} & \textbf{Metrics} & \textbf{AU 1}  & \textbf{AU 2}  & \textbf{AU 4}  & \textbf{AU 5}  & \textbf{AU 6}  & \textbf{AU 12}  & \textbf{AU 25}  & \textbf{AU 26}  \\
        \noalign{\hrule height 0.5pt}
            \multirow{2}{*}{\textbf{ResNet-7}} & AUC  & 0.92 & 0.93 & 0.91 & 0.91 & 0.96 & 0.97 & 0.97 & 0.91 \\
            & EER [\%] & 15.25 & 14.21 & 16.20 & 13.58 & 10.05 & 8.42 & 7.39 & 16.32 \\
        \noalign{\hrule height 0.5pt}
            \multirow{2}{*}{\textbf{VGG-8}} & AUC & 0.89 & 0.87 & 0.89 & 0.90 & 0.96 & 0.96 & 0.96 & 0.90 \\
            & EER [\%] & 16.59 & 16.08 & 16.88 & 14.87 & 9.51 & 8.11 & 7.83 & 16.55 \\
        \noalign{\hrule height 1pt}
    \end{tabular}%
    }
\label{tab:CNN-Tiny}
\end{table}

\begin{figure}[t]
    \centering
    \includegraphics[width=0.8\columnwidth]{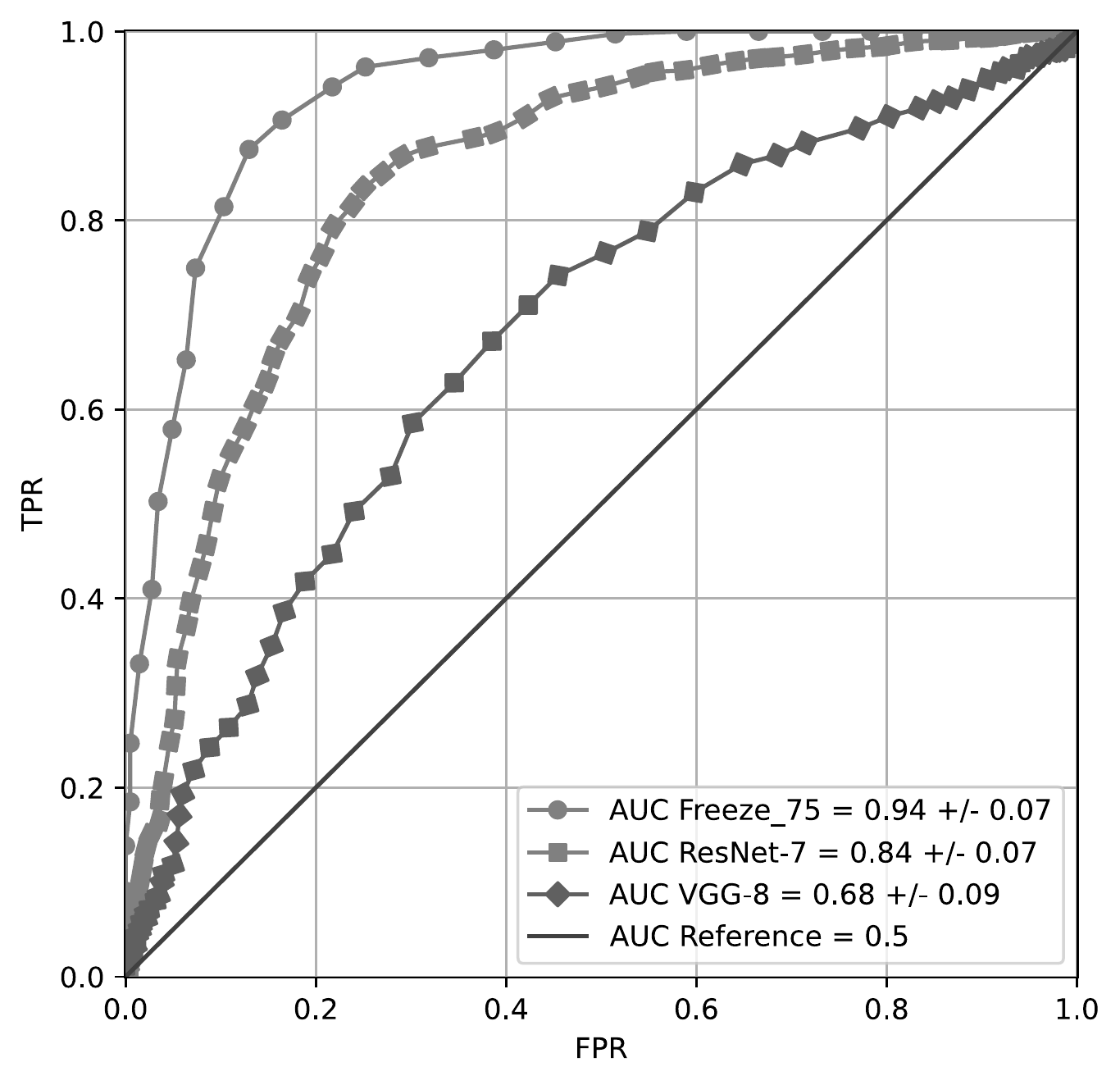}
    \caption{Comparison between PD classification ROC curves obtained using the NOnAOffN sequence in the Freeze 75, ResNet-7 and VGG-8.}
    \label{fig:Vs_Emotion}
\end{figure}

\begin{table}[t]
    \caption{PD classification results using the VGG-8 model.}
    \centering
    \resizebox{\columnwidth}{!}
    {%
    \begin{tabular}{llllll}
    \noalign{\hrule height 1pt}
    \textbf{Sequence} & \textbf{Kernel*} & \textbf{Acc[\%] }   & \textbf{Sens[\%]} & \textbf{Spec[\%]}  & \textbf{F1[\%]}      \\
        \noalign{\hrule height 0.5pt}
        NOnA & $C$=1e+01; $\gamma$=1e-02 	 & 	58.3 $\pm$ 3.7 	 & 	 94.6 $\pm$ 4.8 	 & 	 14.1 $\pm$ 6.3 	 & 	 24.0 $\pm$ 9.8\\
        AOffN &$C$=1e+01; $\gamma$=1e-03 	 & 	65.6 $\pm$ 8.6 	 & 	 80.6 $\pm$ 8.0 	 & 	 47.6 $\pm$ 16.4 	 & 	 58.1 $\pm$ 12.9 \\
        NOnAOffN & $C$=1e+01; $\gamma$=1e-04 	 & 	62.7 $\pm$ 8.3 	 & 	 66.4 $\pm$ 10.0 	 & 	 58.2 $\pm$ 13.1 	 & 	 60.9 $\pm$ 8.6 \\
        \noalign{\hrule height 0.5pt}
        NOnA & $C$=1e-02 	 & 	67.4 $\pm$ 8.3 	 & 	 72.4 $\pm$ 9.4 	 & 	 61.3 $\pm$ 9.8 	 & 	 66.0 $\pm$ 8.2 \\
        \textbf{AOffN} & \textbf{$C$=1e-02} 	 & 	\textbf{67.6 $\pm$ 5.8} 	 & 	 \textbf{70.6 $\pm$ 7.4} 	 & 	 \textbf{63.9 $\pm$ 13.5} 	 & 	 \textbf{65.9 $\pm$ 7.3} \\
        NOnAOffN & $C$=1e-02 	 & 	64.9 $\pm$ 7.7 	 & 	 71.0 $\pm$ 4.5 	 & 	 57.7 $\pm$ 16.1 	 & 	 62.2 $\pm$ 11.0 \\
        \noalign{\hrule height 1pt}
        \multicolumn{6}{l}{First three rows: Gaussian kernel. Last three rows: Linear kernel.}\\
        \multicolumn{6}{l}{*Column with optimal hyper-parameters.}
    \end{tabular}%
    }
    \label{tab:EmotionVGG}
\end{table}

\begin{table}[t]
    \caption{PD classification results using the ResNet-7 model.}
    \centering
    \resizebox{\columnwidth}{!}
    {%
    \begin{tabular}{llllll}
        \noalign{\hrule height 1pt}
        \textbf{Sequence} & \textbf{Kernel*} & \textbf{Acc[\%] }   & \textbf{Sens[\%]} & \textbf{Spec[\%]}  & \textbf{F1[\%]}      \\
        \noalign{\hrule height 0.5pt}
        NOnA & $C$=1e+03; $\gamma$=1e-04 	 & 	73.0 $\pm$ 9.5 	 & 	 75.9 $\pm$ 18.7 	 & 	 69.7 $\pm$ 17.8 	 & 	 68.9 $\pm$ 12.3  \\
        AOffN & $C$=1e+01; $\gamma$=1e-02 	 & 	73.4 $\pm$ 9.9 	 & 	 81.7 $\pm$ 15.6 	 & 	 63.6 $\pm$ 8.9 	 & 	 70.5 $\pm$ 9.5  \\
        \textbf{NOnAOffN} & \textbf{$C$=1e+03; $\gamma$=1e-04} 	 & 	\textbf{78.8 $\pm$ 6.4} 	 & 	 \textbf{79.3 $\pm$ 9.8} 	 & 	 \textbf{78.2 $\pm$ 12.8} 	 & 	 \textbf{77.6 $\pm$ 6.7}  \\
        \noalign{\hrule height 0.5pt}
        NOnA & $C$=1e-02 	 & 	74.1 $\pm$ 6.9 	 & 	 82.2 $\pm$ 19.4 	 & 	 64.5 $\pm$ 11.4 	 & 	 69.3 $\pm$ 6.1  \\
        AOffN & $C$=1e-02 	 & 	72.4 $\pm$ 10.8 	 & 	 84.2 $\pm$ 16.5 	 & 	 58.2 $\pm$ 8.6 	 & 	 68.1 $\pm$ 9.6  \\
        NOnAOffN & $C$=1e-01 	 & 	78.3 $\pm$ 7.3 	 & 	 80.1 $\pm$ 10.6 	 & 	 76.2 $\pm$ 10.1 	 & 	 77.3 $\pm$ 7.4  \\
        \noalign{\hrule height 1pt}
        \multicolumn{6}{l}{First three rows: Gaussian kernel. Last three rows: Linear kernel.}\\
        \multicolumn{6}{l}{*Column with optimal hyper-parameters.}
    \end{tabular}%
    }
    \label{tab:EmotionTiny}
\end{table}

Table~\ref{tab:EmotionVGG} and Table~\ref{tab:EmotionTiny} show
the results obtained when the aforementioned models, created with the
reduced architectures, are used to discriminate between PD patients and
healthy subjects.
Note that no additional training is performed with data from
Parkinson's disease patients.
The best results are obtained when the ResNet-7 architecture is considered
with features extracted from the NOnAOffN sequence. 
Although $78$.$3\%$ could be considered a good accuracy, it is still far 
from the best result obtained with the ResNet50 Freeze 75 model ($87$.$3\%$ in
Table~\ref{tab:Emotion75}), indicating that the FAU domain is missing certain features present in the Face Recognition domain.

Figure~\ref{fig:Vs_Emotion} shows three ROC curves where results with
Freeze 75, ResNet-7, and VGG-8 are compared.
The superiority of the Freeze 75 model is clearly observed, supporting the
advantages of initializing the models using the Face Recognition domain.

\subsection{Experiment 3: Parkinson Domain (PD Detection)}

The triplet loss function is explored for learning in this experiment with the aim to
evaluate whether the classification performance of PD patients vs. Healthy Control (HC) subjects can be improved with respect to previous experiments.
The triplet loss function modifies the original representation space
such that the inter-class separability is increased while the intra-class 
separability is reduced. The modified feature vectors are called
\emph{embedded vectors}.

\subsubsection{Triplet Loss in Face Recognition models adapted to the Affective domain}

The Freeze 75 and Freeze 50 models are trained with the triplet loss function 
strategy and two new models are obtained, namely Triplet 75 and Triplet 50,
respectively. The FacePark-GITA database is divided into a 5-fold partition 
for the training of each Triplet model and the SVM classifier.
The classification results obtained when using the embedded vectors are
shown in Table~\ref{tab:Triplet75} for the Triplet 75 model, and in
Table~\ref{tab:Triplet50} for the Triplet 50 model.

\begin{table}[t]
    \caption{PD classification results of classification with the Triplet 75 model.}
    \centering
    \resizebox{.48\textwidth}{!}
    {%
    \begin{tabular}{llllll}
        \noalign{\hrule height 1pt}
        \textbf{Sequence} &\textbf{Kernel*} & \textbf{Acc[\%] }   & \textbf{Sens[\%]} & \textbf{Spec[\%]}  & \textbf{F1[\%]}   \\
        \noalign{\hrule height 0.5pt}
        NOnA & $C$=1e+01; $\gamma$=1e-04 	 & 	85.2 $\pm$ 7.4 	 & 	 87.6 $\pm$ 5.8 	 & 	 82.5 $\pm$ 12.6 	 & 	 84.5 $\pm$ 8.2 \\
        AOffN & $C$=1e+01; $\gamma$=1e-04 	 & 	86.0 $\pm$ 6.1 	 & 	 91.4 $\pm$ 6.9 	 & 	 79.5 $\pm$ 7.1 	 & 	 84.9 $\pm$ 6.2\\
        NOnAOffN & $C$=1e+01; $\gamma$=1e-04 	 & 	86.0 $\pm$ 9.0 	 & 	 92.1 $\pm$ 6.9 	 & 	 78.7 $\pm$ 13.4 	 & 	 84.5 $\pm$ 10.1 \\
        \noalign{\hrule height 0.5pt}
        NOnA & $C$=1e-01 	 & 	84.4 $\pm$ 6.6 	 & 	 87.4 $\pm$ 4.4 	 & 	 80.9 $\pm$ 13.3 	 & 	 83.4 $\pm$ 7.6 \\
        AOffN & $C$=1e-01 	 & 	85.0 $\pm$ 5.9 	 & 	 90.3 $\pm$ 6.4 	 & 	 78.7 $\pm$ 7.1 	 & 	 84.0 $\pm$ 6.1 \\
        \textbf{NOnAOffN} & \textbf{$C$=1e-01} 	 & 	\textbf{86.1 $\pm$ 9.6} 	 & 	 \textbf{91.4 $\pm$ 7.5} 	 & 	 \textbf{79.9 $\pm$ 13.5} 	 & 	 \textbf{85.0 $\pm$ 10.5} \\
        \noalign{\hrule height 1pt}
        \multicolumn{6}{l}{First three rows: Gaussian kernel. Last three rows: Linear kernel.}\\
        \multicolumn{6}{l}{*Column with optimal hyper-parameters.}
    \end{tabular}%
    }
    \label{tab:Triplet75}
\end{table}

\begin{table}[t]
    \caption{PD classification results of classification with the Triplet 50 model.}
    \centering
    \resizebox{.48\textwidth}{!}
    {%
    \begin{tabular}{llllll}
        \noalign{\hrule height 1pt}
        \textbf{Sequence} & \textbf{Kernel*} & \textbf{Acc[\%] }   & \textbf{Sens[\%]} & \textbf{Spec[\%]}  & \textbf{F1[\%]}   \\
        \noalign{\hrule height 0.5pt}
        NOnA & $C$=1e+01; $\gamma$=1e-04 	 & 	78.9 $\pm$ 5.5 	 & 	 84.3 $\pm$ 10.9 	 & 	 72.4 $\pm$ 11.3 	 & 	 76.7 $\pm$ 6.1 \\
        AOffN & $C$=1e+03; $\gamma$=1e-04 	 & 	73.2 $\pm$ 8.7 	 & 	 69.1 $\pm$ 16.9 	 & 	 78.3 $\pm$ 4.0 	 & 	 72.2 $\pm$ 8.3 \\
        NOnAOffN & $C$=1e+02; $\gamma$=1e-04 	 & 	75.8 $\pm$ 11.8 	 & 	 77.4 $\pm$ 15.5 	 & 	 74.3 $\pm$ 16.2 	 & 	 74.2 $\pm$ 12.5 \\
        \noalign{\hrule height 0.5pt}
        \textbf{NOnA} & \textbf{$C$=1e-01} 	 & 	\textbf{80.7 $\pm$ 6.6} 	 & 	 \textbf{86.4 $\pm$ 13.2} 	 & 	 \textbf{73.9 $\pm$ 11.8} 	 & 	 \textbf{78.1 $\pm$ 7.4} \\
        AOffN & $C$=1e-01 	 & 	76.3 $\pm$ 8.7 	 & 	 79.1 $\pm$ 17.4 	 & 	 73.3 $\pm$ 7.4 	 & 	 74.5 $\pm$ 8.6 \\
        NOnAOffN & $C$=1e-01 	 & 	77.1 $\pm$ 10.2 	 & 	 83.0 $\pm$ 10.7 	 & 	 69.9 $\pm$ 19.8 	 & 	 73.9 $\pm$ 13.2 \\
        \noalign{\hrule height 1pt}
        \multicolumn{6}{l}{First three rows: Gaussian kernel. Last three rows: Linear kernel.}\\
        \multicolumn{6}{l}{*Column with optimal hyper-parameters.}
    \end{tabular}%
    }
    \label{tab:Triplet50}
\end{table}

Note that the Triplet 75 model exhibits better accuracy ($86$.$0\%$) than
the Triplet 50 ($80$.$7\%$).
Since the best accuracies in the previous experiments 
with the Freeze 75 and Freeze 50 models were $87$.$3\%$ and $83$.$1\%$, these
new results obtained with the triplet loss strategy likely indicate
that the embedding approach does not provide advantages over the use
of transfer learning and freezing of layers. 
This observation is also supported in the fact that the number of parameters
to be optimized has not been reduced, so in principle, there is no
reason for using the triplet loss function in these two scenarios.
\begin{figure*}[t]
\centering

    \begin{subfigure}{0.3\textwidth}
      \centering
      \includegraphics[width=1\linewidth]{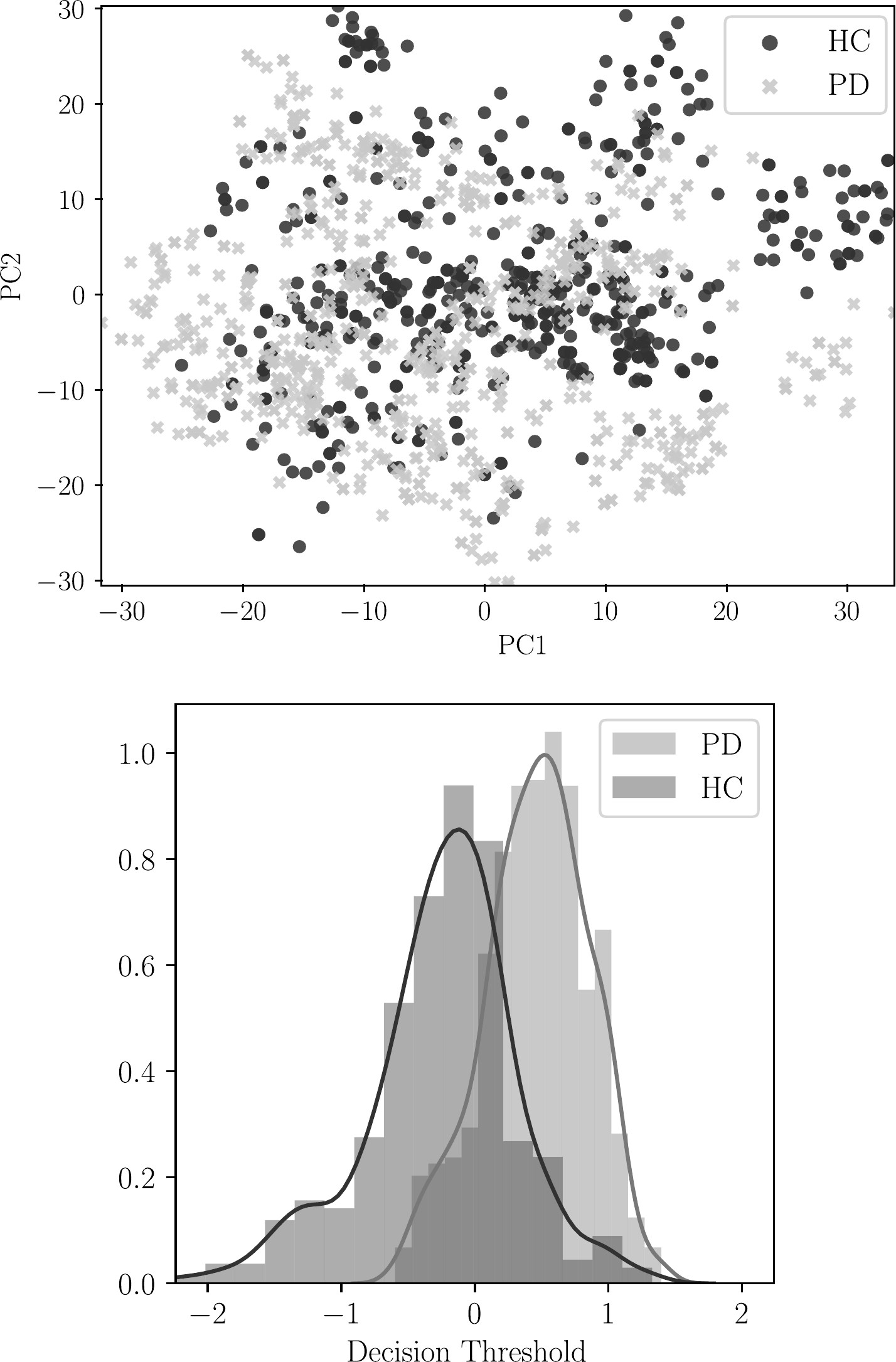}  
      \caption{ResNet50 ($\textbf{x}_{\textrm{FR}}$)}
    \end{subfigure}
    \begin{subfigure}{0.3\textwidth}
      \centering
      \includegraphics[width=1\linewidth]{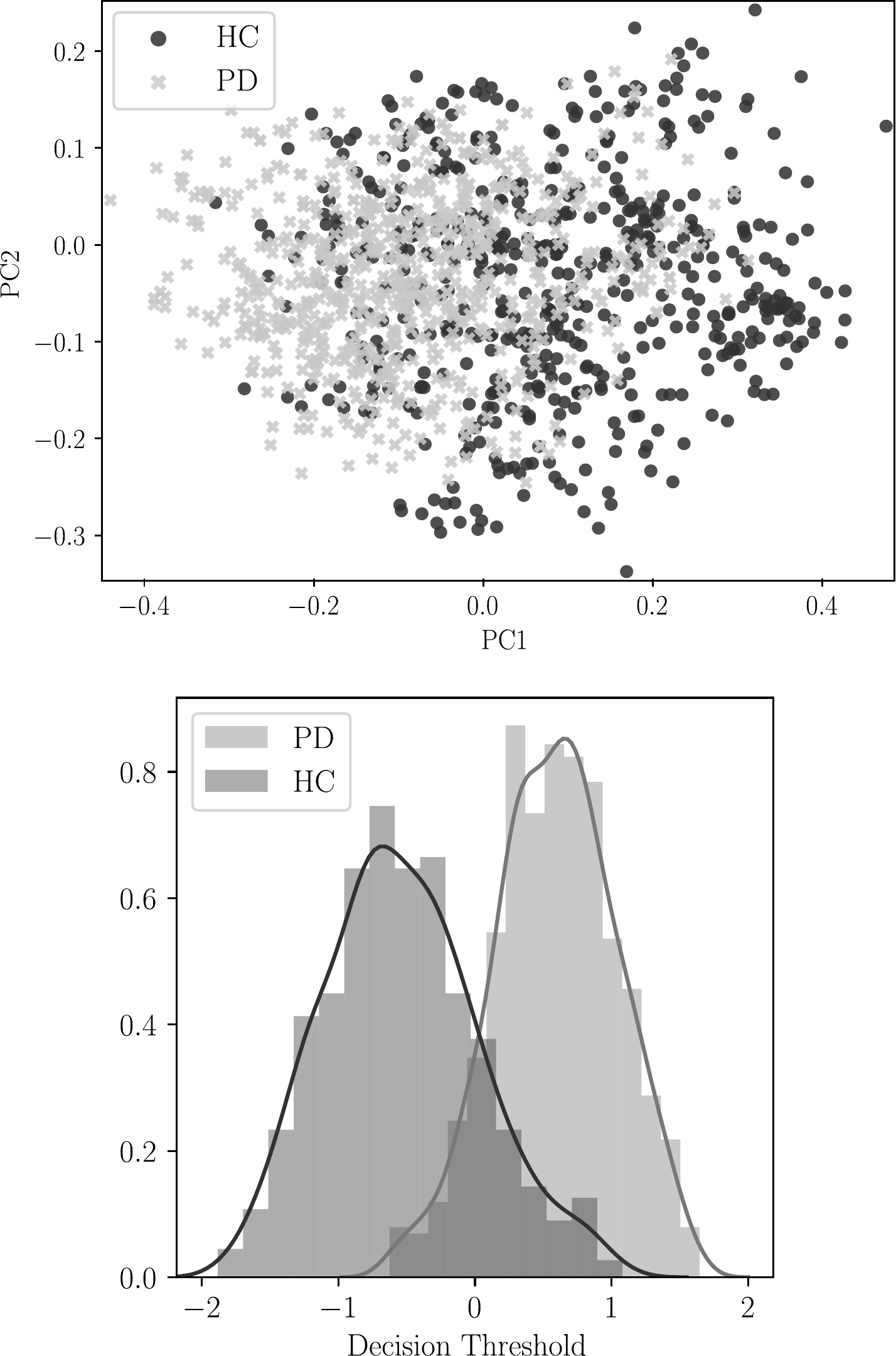}  
      \caption{ResNet50+FAU ($\textbf{x}_{\textrm{AF}}$)}
    \end{subfigure}
    \begin{subfigure}{0.3\textwidth}
      \centering
      \includegraphics[width=1\linewidth]{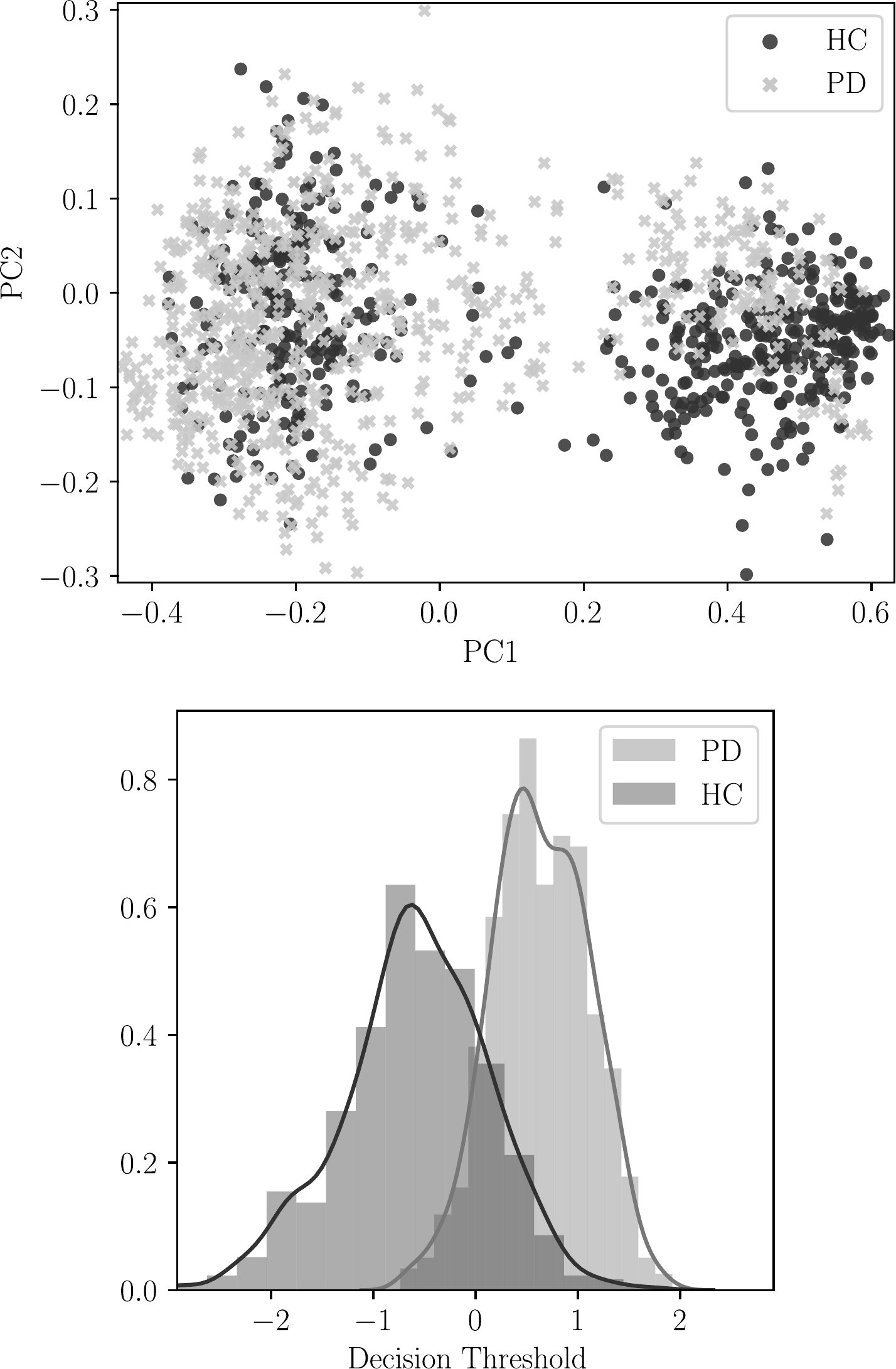}  
      \caption{Triplet-ResNet7 ($\textbf{x}_{\textrm{PD}}$)}
    \end{subfigure}

    \caption{(Up) Principal components spaces generated from the features of the
    different models and (Bottom) score distributions of PD patients and Healthy Control (HC) subjects obtained by the SVM classifier.}
    \label{fig:feature_spaces}
\end{figure*}

\subsubsection{Triplet Loss in FAU detection trained from scratch}

In this experiment the VGG-8 and ResNet-7 models are retrained 
considering the triplet loss function, creating two new models, 
namely Triplet-VGG8 and Triplet-ResNet7, respectively. 
These new models are used to extract embedded vectors for further 
classification between PD patients and healthy subjects. 
The results obtained with the Triplet-VGG8 and Triplet-ResNet7 embedded
vectors are shown in 
Table~\ref{tab:Triplet-VGG8} and Table~\ref{tab:Triplet-ResNet7}, respectively.

\begin{table}[t]
    \caption{PD classification results using the Triplet-VGG8 model.}
    \centering
    \resizebox{\columnwidth}{!}
    {%
    \begin{tabular}{llllll}
        \noalign{\hrule height 1pt}
        \textbf{Sequence} & \textbf{Kernel*} & \textbf{Acc[\%] }   & \textbf{Sens[\%]} & \textbf{Spec[\%]}  & \textbf{F1[\%]}   \\
        \noalign{\hrule height 0.5pt}
        NOnA & $C$=1e+01; $\gamma$=1e-04 	 & 	71.2 $\pm$ 8.8 	 & 	 76.4 $\pm$ 14.0 	 & 	 64.9 $\pm$ 12.8 	 & 	 68.7 $\pm$ 8.2 \\
        AOffN & $C$=1e+03; $\gamma$=1e-03 	 & 	69.9 $\pm$ 9.6 	 & 	 67.4 $\pm$ 8.2 	 & 	 72.9 $\pm$ 13.1 	 & 	 69.8 $\pm$ 9.6 \\
        NOnAOffN & $C$=1e+00; $\gamma$=1e-03 	 & 	66.0 $\pm$ 8.4 	 & 	 79.0 $\pm$ 10.5 	 & 	 50.7 $\pm$ 21.0 	 & 	 58.2 $\pm$ 14.9 \\
        \noalign{\hrule height 0.5pt}
        \textbf{NOnA} & \textbf{$C$=1e-02} 	 & 	\textbf{72.7 $\pm$ 7.2 }	 & 	\textbf{ 80.8 $\pm$ 13.4} 	 & 	 \textbf{62.6 $\pm$ 11.5} 	 & 	 \textbf{69.1 $\pm$ 7.9} \\
        AOffN & $C$=1e+01 	 & 	70.3 $\pm$ 7.0 	 & 	 74.9 $\pm$ 9.4 	 & 	 64.8 $\pm$ 13.2 	 & 	 68.3 $\pm$ 7.8 \\
        NOnAOffN & $C$=1e+01 	 & 	65.3 $\pm$ 5.1 	 & 	 65.0 $\pm$ 3.9 	 & 	 65.4 $\pm$ 13.7 	 & 	 64.1 $\pm$ 6.8 \\
        \noalign{\hrule height 1pt}
        \multicolumn{6}{l}{First three rows: Gaussian kernel. Last three rows: Linear kernel.}\\
        \multicolumn{6}{l}{*Column with optimal hyper-parameters.}
    \end{tabular}%
    }
    \label{tab:Triplet-VGG8}
\end{table}

\begin{table}[t]
    \caption{PD classification results using the Triplet-ResNet7 model.}
    \centering
    \resizebox{\columnwidth}{!}
    {%
    \begin{tabular}{llllll}
        \noalign{\hrule height 1pt}
        \textbf{Sequence} & \textbf{Kernel*} & \textbf{Acc[\%] }   & \textbf{Sens[\%]} & \textbf{Spec[\%]}  & \textbf{F1[\%]}   \\
        \noalign{\hrule height 0.5pt}
        NOnA & $C$=1e+03; $\gamma$=1e-04 	 & 	82.1 $\pm$ 8.8 	 & 	 87.2 $\pm$ 7.4 	 & 	 76.0 $\pm$ 14.3 	 & 	 80.5 $\pm$ 10.1 \\
        AOffN & $C$=1e+02; $\gamma$=1e-03 	 & 	78.2 $\pm$ 12.9 	 & 	 79.6 $\pm$ 13.6 	 & 	 76.3 $\pm$ 16.3 	 & 	 77.3 $\pm$ 13.0 \\
        NOnAOffN & $C$=1e-01; $\gamma$=1e-03 	 & 	69.9 $\pm$ 10.8 	 & 	 82.8 $\pm$ 15.2 	 & 	 54.7 $\pm$ 22.0 	 & 	 61.8 $\pm$ 17.9 \\
        \noalign{\hrule height 0.5pt}
        \textbf{NOnA} & \textbf{$C$=1e-01} 	 & 	\textbf{82.4 $\pm$ 8.5} 	 & 	 \textbf{89.2 $\pm$ 5.9}	 & 	 \textbf{74.1 $\pm$ 12.6} 	 & 	 \textbf{80.7 $\pm$ 9.7} \\
        AOffN & $C$=1e-01 	 & 	76.2 $\pm$ 11.0 	 & 	 78.9 $\pm$ 12.5 	 & 	 72.8 $\pm$ 12.7 	 & 	 75.3 $\pm$ 11.0 \\
        NOnAOffN & $C$=1e-02 	 & 	79.6 $\pm$ 5.4 	 & 	 89.0 $\pm$ 11.0 	 & 	 68.6 $\pm$ 10.3 	 & 	 76.5 $\pm$ 5.1 \\
        \noalign{\hrule height 1pt}
        \multicolumn{6}{l}{First three rows: Gaussian kernel. Last three rows: Linear kernel.}\\
        \multicolumn{6}{l}{*Column with optimal hyper-parameters.}
    \end{tabular}%
    }
    \label{tab:Triplet-ResNet7}
\end{table}

Note that there is an improvement in both models compared to those based on
VGG-8 and ResNet-7 where the triplet loss function was not applied.
In the first case the improvement is around $5$.$1\%$ (from $67$.$6\%$ to $72$.$7\%$) and
in the second case is around $3$.$6\%$ (from $78$.$8\%$ to $82$.$4\%$). These results partially validates our third hypothesis (\textit{H3}) indicating that loss functions designed to learn from the PD domain serve to improve the performance of PD classification. 
It is not only interesting to highlight the improvement achieved when using
the triplet loss function, but also to note that the best result obtained
with the Triplet-ResNet7 model is competitive compared to the best accuracy
previously obtained with the Freeze 75 model. Although the accuracy in the 
second one is $4$.$9\%$ above the first one, Freeze 75 requires 
$17$,$815$,$520$ more parameters to be optimized than Triplet-ResNet7, which
might indicate a better generalization capability. Further experiments
with additional data are required to validate this hypothesis.

PCA is now used to create a 2D representation 
of the feature spaces learned in previous experiments.
Figure~\ref{fig:feature_spaces} shows the feature spaces and the distribution of the 
classification scores. The figure shows a superior discrimination capability of the $\textbf{x}_{\textrm{AF}}$ feature space (ResNet50 adapted to the FAU domain). The representation obtained by the Triplet-ResNet7 model shows a larger margin between classes but the misclassification errors decrease the performance.

\subsection{Experiment 4: Parkinson Domain (PD Impairment Estimation)}

\begin{figure}[t]
    \centering
    \includegraphics[width=0.8\columnwidth]{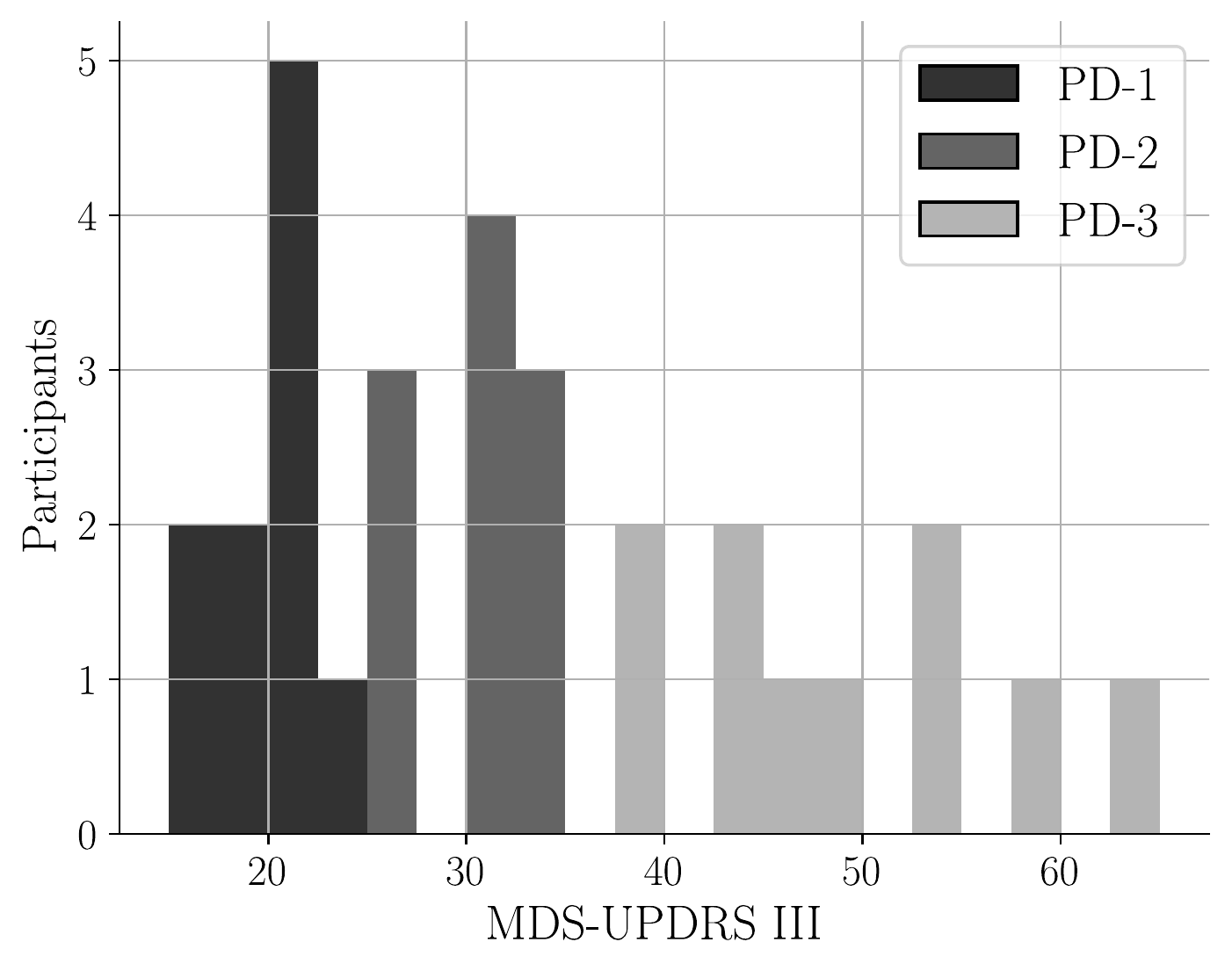}
    \caption{Distribution of the neurological state of the Parkinson's patients according to their score in the MDS-UPDRS-III scale.}
    \label{fig:UPDRS2Class}
\end{figure}

Given the promising results obtained with the above presented experiments in the automatic discrimination between PD patients and healthy subjects, especially with the Freeze~75 and the Triplet-ResNet7 models,
with accuracies of $87$.$3\%$ and $82$.$4\%$, respectively, 
we want to evaluate in this section the suitability of those models to 
discriminate between three different degrees of impairment:
mild (PD-1), intermediate (PD-2), and severe (PD-3).
These three groups are defined considering the scores of the MDS-UPDRS-III 
provided by the expert neurologist.
The mild group includes patients with scores in the range from 0 to 23, 
the intermediate group is defined for patients with scores between 23 and 33, 
and the severe group for patients with scores greater than 33. Figure~\ref{fig:UPDRS2Class} shows the distribution of the MDS-UPDRS-III
scores for the three groups of patients. 

The tri-class classification experiments are performed considering the
feature vectors extracted with the Freeze 75 model on the NOnAOffN sequence, and
the Triplet-ResNet7 model on the NOnA sequence.
Optimization of hyper-parameters is performed as indicated in
Section~\ref{sec:classification}.
The confusion matrices and results obtained with the Freeze 75 and 
the Triplet-ResNet7 models
are shown in Table~\ref{tab:MultiClassWOHC} and 
Table~\ref{tab:MultiClassWOHCTiny}, respectively.
Values of accuracy, F1 score, and $\kappa$ index are included in the
bottom part of each table.
\begin{table}[t]
    \caption{Confusion matrix for the classification of PD patients
    with different degree of impairments using the Freeze 75 model
    with feature vectors extracted from the NOnAOffN sequence.}
    \centering
    \resizebox{0.55\columnwidth}{!}
    {%
        \begin{tabular}{lllll}
        \noalign{\hrule height 1pt}
         & \multicolumn{3}{c}{SVM: $C$=1e-03} &\\
         \cline{2-4} 
         & PD-1 & PD-2 & PD-3 & \\ 
         \noalign{\hrule height 1pt}
        PD-1 & 45.80 & 35.30 & 18.90 & \\
        PD-2 & 33.47 & 42.26 & 24.27 & \\
        PD-3 & 15.45 & 39.49 & 45.06 & \\ 
        \noalign{\hrule height 1pt} 
        \multicolumn{5}{l}{SVM: Acc= 44\%, F1= 0.45, $\kappa$= 0.17.}
        \end{tabular}%
    }
    \label{tab:MultiClassWOHC}
\end{table}
\begin{table}[t]
    \caption{Confusion matrix for the classification of PD patients
    with different degree of impairments using the Triplet-ResNet7 model
    with feature vectors extracted from the NOnA sequence.}
    \centering
    \resizebox{0.55\columnwidth}{!}
    {%
        \begin{tabular}{lllll}
        \noalign{\hrule height 1pt}
         & \multicolumn{3}{c}{SVM: $C$=1e-03} & \\
         \cline{2-4}  
         & PD-1 & PD-2 & PD-3 & \\ 
         \noalign{\hrule height 1pt}
        PD-1 & 30.07 & 39.60 & 30.33 & \\
        PD-2 & 34.60 & 20.53 & 44.87 & \\
        PD-3 & 10.28 & 18.62 & 71.10 & \\ 
        \noalign{\hrule height 1pt} 
        \multicolumn{5}{l}{SVM:  Acc= 40\%, F1= 0.38, $\kappa$= 0.11.}
        \end{tabular}%
    }
    \label{tab:MultiClassWOHCTiny}
\end{table}
These results show that the Freeze 75 model is better than the Triplet-ResNet7.
There is a difference of $4\%$ points in the accuracy, and $0.07$ in the 
F1 score when using the SVM. 

\begin{figure}[t] 
    \centering
    \includegraphics[width=0.9\columnwidth]{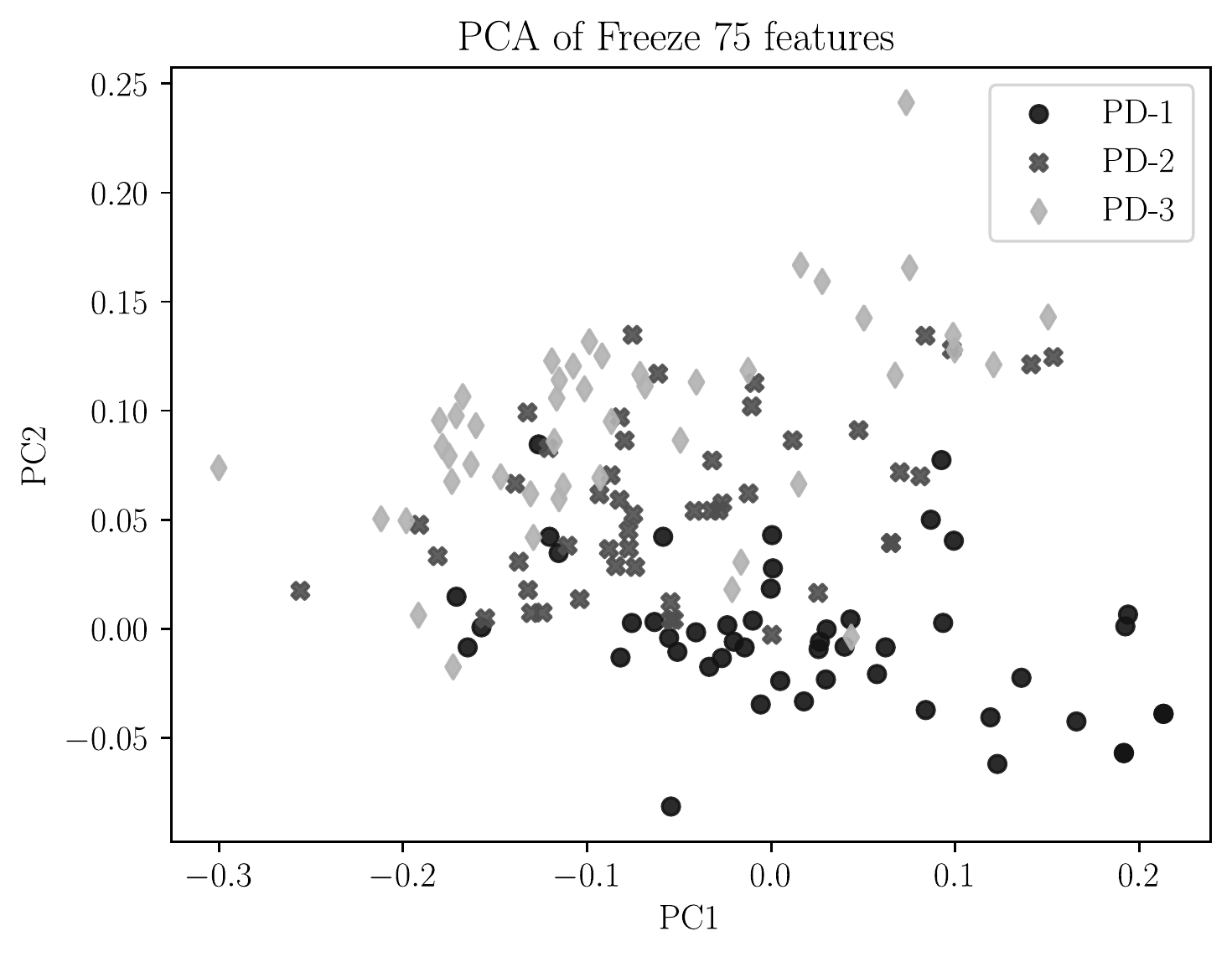}
    \caption{Principal component space created with feature vectors of 
    the Freeze 75 model of PD patients of three groups with different degree of impairment.}
    \label{fig:PCA_Freeze75}
\end{figure}

The accuracies obtained are  around $43\%$ for a problem with three classes ($33\%$ random chance). These results are far to be optimal but suggest there is certain useful information in the models that can help to estimate the degree of impairment (our fourth hypothesis \textit{H4}).  Figure~\ref{fig:PCA_Freeze75} shows the projected space. Note that 
there is a relatively clear
separability between mild and severe patients.
In the middle between these two groups there are samples of the
intermediate group, which are not accurately classified but clearly appear 
in between, which likely indicates that the proposed approach is able to
find a trend regarding the neurological state of the patients.

\section{Discussion and Conclusion}

This study presents a novel approach where deep learning methods are used to 
model hypomimia in PD patients. 
Videos with the face of people while evoking emotions are considered for the study.
Frames of the recorded videos are segmented in different stages during the 
production of evoked emotions: neutral, onset-transition, apex, offset-transition, and neutral.
This approach exhibits improvements of up to 5.5\% in accuracy
(from 72.9\% to 78.4\%) with respect to 
classical approaches where single frames are considered. 
These results suggest that dynamics information is more suitable to model hypomimia
in PD patients. We are aware of the fact that the presented approach does not completely exploit the video dynamics; however, the incorporation of frames in different stages during the production of emotions shows to be a good and computationally affordable approach.

Later, information from the Affective domain is incorporated in the model by means of transfer learning methods. Transfer learning was performed considering the complete architecture of a base model previously trained with massive data and then freezing some layers to fine-tune the remaining layers with the smaller emotion data. Results freezing 75\% and 50\% of the layers are reported.
The results show that the Affective domain adaptation provides an improvement of 8.9\%,
from 78.4\% to 87.3\% of accuracy in PD detection.
These results confirm that domain adaptation via transfer learning methods 
is a good strategy to model hypomimia in PD patients. Considering the good results
and also the fact that only up to four images per participant are considered in the
experiments, we believe that this work is a step forward in the development of 
inexpensive computational systems suitable to model and quantify problems of 
PD patients to express emotions. 

With the aim of finding lighter approaches suitable to be used in portable devices,
other experiments with reduced architectures like VGG-8 and ResNet7 were also addressed, 
however, the results were not as good as before, the maximum achieved accuracies in this case were 67.6\% and 78.8\%, respectively.
These results were further improved up to 72.7\% and 82.4\%
when the triplet loss strategy is considered for training the VGG-8 and ResNet7 models, 
respectively. 

Finally, the neurological state of the patients was evaluated considering the best approach found in the classification experiments.
The patients were grouped into three groups according to their MDS-UPDRS-III scores and a tri-class classification strategy showed a maximum classification accuracy of 44\% (F1=0.45).

Future work includes more sophisticate methods to integrate the information provided by the full video sequences, including video tracking of facial features. We will also investigate multiple classifier approaches to combine the information provided by face videos for PD detection and PD impairment estimation with other sources of information \cite{2018_INFFUS_MCSreview1_Fierrez}, like speech, gait, handwriting \cite{2020_COGN_HandwTrends_Faundez}, and human-computer interaction signals \cite{2020_CDS_HCIsmart_Acien}.

\section*{Acknowledgements}
The authors would like to thank the patients of the Parkinson's
Foundation in Medell\'in, Colombia (Fundalianza\footnote{https://www.fundalianzaparkinson.org/}) for 
their cooperation during the development of this study.
The study was partially funded by CODI at Universidad de Antioquia
grants \# PRG2017-15530 \& PRG2015-7683, and also by the 
ASC strategy of the Ministry of Science in Colombia. Authors J.F. and A.M. are funded by projects BIBECA (RTI2018-101248-B-I00 MINECO/FEDER) and IDEA-FAST (H2020-IMI2-2018-15-853981).
We would also like to thank Professor Jesús Francisco Vargas Bonilla for his contributions in the preliminary experiments of this study.

\section*{COMPLIANCE WITH ETHICAL STANDARDS}

\subsection*{Ethical approval}

All of the signals considered in this work were collected in compliance
with the Helsinki Declaration and the procedure was approved by the Ethics
Committee at the University of Antioquia in Medellín, Colombia. Written
informed consent was signed by each participant.


\printbibliography

@INPROCEEDINGS{2019_FG_PDhandw_Castrillon,
author = {R. Castrillon and A. Acien and J.R. Orozco-Arroyave and A. Morales and J.F. Vargas and R. Vera-Rodriguez and J. Fierrez and J. Ortega-Garcia and A. Villegas},
booktitle = {IEEE Intl. Conf. on Automatic Face and Gesture Recognition (FG)},
month = {May},
title = {Characterization of the Handwriting Skills as a Biomarker for Parkinson Disease},
year = {2019},
}

@ARTICLE{2020_COGN_HandwTrends_Faundez,
author={Faundez-Zanuy, M. and Fierrez, J. and Ferrer, M. A. and Diaz, M. and Tolosana, R. and Plamondon, R.},
journal = {Cognitive Computation},
month = {August},
title = {Handwriting Biometrics: Applications and Future Trends in e-Security and e-Health},
year = {2020},
}

@INPROCEEDINGS{2021_ICPR_Emotional_Pena,
author={Peña, A. and Fierrez, J. and Lapedriza, A. and Morales, A.},
booktitle = {IAPR Intl. Conf. on Pattern Recognition (ICPR)},
month = {January},
title = {Learning Emotional-Blinded Face Representations},
year = {2021},
}

@ARTICLE{2013PTomeFSI_FacialRegions,
author = {P. Tome and J. Fierrez and R. Vera-Rodriguez and D. Ramos},
journal = {Forensic Science International},
number = {233},
pages = {75-83},
title = {Identification using Face Regions: Application and Assessment in Forensic Scenarios},
year = {2013},
}

@ARTICLE{2018_IntelligentSystems_icb-rw,
author = {E. Gonzalez-Sosa and R. Vera-Rodriguez and J. Fierrez and J. Ortega-Garcia},
journal = {IEEE Intelligent Systems},
month = {May},
number = {3},
pages = {60-63},
title = {Exploring Facial Regions in Unconstrained Scenarios: Experience on ICB-RW},
volume = {33},
year = {2018},
}

@ARTICLE{2018_TIFS_SoftWildAnno_Sosa,
author={Gonzalez-Sosa, E. and Fierrez, J. and Vera-Rodriguez, R. and Alonso-Fernandez, F.},
journal = {IEEE Trans. on Information Forensics and Security},
month = {August},
number = {8},
pages = {2001-2014},
title = {Facial Soft Biometrics for Recognition in the Wild: Recent Works, Annotation and COTS Evaluation},
volume = {13},
year = {2018},
}

@ARTICLE{2018_INFFUS_MCSreview2_Fierrez,
author={Fierrez, J. and Morales, A. and Vera-Rodriguez, R. and Camacho, D.},
journal = {Information Fusion},
month = {November},
pages = {103-112},
title = {Multiple Classifiers in Biometrics. Part 2: Trends and Challenges},
volume = {44},
year = {2018},
}

@ARTICLE{2018_INFFUS_MCSreview1_Fierrez,
author={Fierrez, J. and Morales, A. and Vera-Rodriguez, R. and Camacho, D.},
journal = {Information Fusion},
month = {November},
pages = {57-64},
title = {Multiple Classifiers in Biometrics. Part 1: Fundamentals and Review},
volume = {44},
year = {2018},
}

@INPROCEEDINGS{2020_CDS_HCIsmart_Acien,
author = {Alejandro Acien and Aythami Morales and Ruben Vera-Rodriguez and Julian Fierrez and Oscar Delgado},
booktitle = {IEEE Conf. on Computers, Software, and Applications (COMPSAC)},
month = {July},
title = {Smartphone Sensors For Modeling Human-Computer Interaction: General Outlook And Research Datasets For User Authentication},
year = {2020},
}

@inproceedings{parkhi2015face,
  title = {{Deep Face Recognition}},
  author = {Parkhi, O. M. and Vedaldi, A. and Zisserman, A. and others},
  booktitle = {British Machine Vision Conference (BMVC)},
  address = {Swansea, UK},
  year = {2015}
}

@article{guillen2018affective,
  title={Affective Robots: Evaluation of Automatic Emotion Recognition Approaches on a Humanoid Robot towards Emotionally Intelligent Machines},
  author={Guill{\'e}n, S. S. and Iacono, L. L. and Meder, C.},
  journal={Energy},
  volume={3643},
  year={2018}
}

@article{pantic2000expert,
  title={Expert system for automatic analysis of facial expressions},
  author={Pantic, M. and Rothkrantz, L. J.},
  journal={Image and Vision Computing},
  volume={18},
  number={11},
  pages={881--905},
  year={2000},
  publisher={Elsevier}
}

@article{li2018deep,
  title={Deep facial expression recognition: A survey},
  author={Li, S. and Deng, W.},
  journal={IEEE Transactions on Affective Computing},
  volume={2010},
  year={2020}
}

@inproceedings{he2016deep,
  title={Deep residual learning for image recognition},
  author={He, K. and Zhang, X. and Ren, S. and Sun, J.},
  booktitle={Proceedings of IEEE Computer Society Conference on Computer Vision and Pattern Recognition},
  pages={770--778},
  year={2016}
}

@misc{ekman1978manual,
  title={Facial Action Coding System: A technique for the measurement of facial movement.},
  author={Friesen, E. and Ekman, P.},
  year={1978},
  city={Palo Alto},
  publisher={CA: Consulting Psychologist Press}
}

@article{bandini2017analysis,
  title={Analysis of facial expressions in Parkinson's disease through video-based automatic methods},
  author={Bandini, A. and Orlandi, S. and Escalante, H. J. and Giovannelli, F. and others},
  journal={Journal of Neuroscience Methods},
  volume={281},
  pages={7--20},
  year={2017},
  publisher={Elsevier}
}

@article{Richardson2000,
  title={Digitizing the moving face during dynamic expressions of emotion},
  author={Richardson, C. and Bowers, D. and Bauer, R. and Heilman, K. M. and Leonard, C. M},
  journal={Neuropsychologia},
  volume={38},
  pages={1026--1037},
  year={2000}
}

@article{Bowers2006,
  title={Faces of emotion in Parkinson's disease: Micro-expressivity and bradykinesia during voluntary facial expressions},
  author={Bowers, D. and Miller, K. and Bosch, W. and Gokcay, D. and Pedraza, O. and Springer, U. and Okun, M.},
  journal={Journal of the International Neuropsychological Society},
  volume={12},
  pages={765-773},
  year={2006}
}

@article{Almutiry2016,
  title={Facial Behaviour Analysis in Parkinson's Disease},
  author={Almutiry, R. and Couth, S. and Poliakoff, E. and Kotz, S. and Silverdale, M. and Cootes, T.},
  journal={Lecture Notes in Computer Science},
  volume={9805},
  pages={329--339},
  year={2016}
}

@article{Gunnery2017,
  title={Mapping spontaneous facial expression in people with Parkinson's 
  disease: A multiple case study design},
  author={Gunnery, S. D. and Naumova, E. N. and Saint-Hilaire, M. and Tickle-Degnen, L.},
  journal={Cogent Psychology},
  volume={4},
  pages={1--15},
  year={2017}
}

@article{cacabelos2017parkinson,
  title={Parkinson’s disease: from pathogenesis to pharmacogenomics},
  author={Cacabelos, R.},
  journal={International Journal of Molecular Sciences},
  volume={18},
  number={3},
  pages={551},
  year={2017},
  publisher={Multidisciplinary Digital Publishing Institute}
}

@article{bologna2013facial,
  title={Facial bradykinesia},
  author={Bologna, M. and Fabbrini, G. and Marsili, L. and Defazio, G. and Thompson, P. D. and Berardelli, A.},
  journal={J Neurol Neurosurg Psychiatry},
  volume={84},
  number={6},
  pages={681--685},
  year={2013},
  publisher={BMJ Publishing Group Ltd}
}

@article{goetz2008movement,
  title={Movement Disorder Society-sponsored revision of the Unified Parkinson's Disease Rating Scale (MDS-UPDRS): scale presentation and clinimetric testing results},
  author={Goetz, C. G. and Tilley, B. C. and Shaftman, S. R. and Stebbins, G. T. and Fahn, S. and Martinez‐Martin, P and others},
  journal={Movement Disorders},
  volume={23},
  number={15},
  pages={2129--2170},
  year={2008},
  publisher={Wiley Online Library}
}

@article{pan2009survey,
  title={A survey on transfer learning},
  author={Pan, S. J. and Yang, Q.},
  journal={IEEE Transactions on Knowledge and Data Engineering},
  volume={22},
  number={10},
  pages={1345--1359},
  year={2009},
  publisher={IEEE}
}

@article{goetz2004movement,
  title={Movement Disorder Society Task Force report on the Hoehn and Yahr staging scale: status and recommendations the Movement Disorder Society Task Force on rating scales for Parkinson's disease},
  author={Goetz, C. G. and Poewe, W. and Rascol, O. and Sampaio, C. and Stebbins, G. T. and Counsell, C. and Giladi, N. and Holloway, R. G. and Moore, C. G. and Wenning, G. K. and others},
  journal={Movement Disorders},
  volume={19},
  number={9},
  pages={1020--1028},
  year={2004},
  publisher={Wiley Online Library}
}

@inproceedings{fabian2016emotionet,
  title={Emotionet: An accurate, real-time algorithm for the automatic annotation of a million facial expressions in the wild},
  author={Fabian Benitez-Quiroz, C. and Srinivasan, R. and Martinez, A. M.},
  booktitle={Proceedings of the IEEE Conference on Computer Vision and Pattern Recognition},
  pages={5562--5570},
  year={2016}
}

@inproceedings{yan2019vargfacenet,
  title={Vargfacenet: An efficient variable group convolutional neural network for lightweight face recognition},
  author={Yan, M. and Zhao, M. and Xu, Z. and Zhang, Q. and Wang, G. and Su, Z.},
  booktitle={Proceedings of the IEEE International Conference on Computer Vision Workshops},
  year={2019}
}

@article{ranjan2018deep,
  title={Deep learning for understanding faces: Machines may be just as good, or better, than humans},
  author={Ranjan, R. and Sankaranarayanan, S. and Bansal, A. and Bodla, N. and Chen, J. C. and others},
  journal={IEEE Signal Processing Magazine},
  volume={35},
  number={1},
  pages={66--83},
  year={2018},
  publisher={IEEE}
}

@inproceedings{cao2018vggface2,
  title={Vggface2: A dataset for recognising faces across pose and age},
  author={Cao, Q. and Shen, L. and  Xie, W. and Parkhi, O. M. and Zisserman, A.},
  booktitle={2018 13th IEEE International Conference on Automatic Face \& Gesture Recognition (FG 2018)},
  pages={67--74},
  year={2018},
  organization={IEEE}
}

@article{ekman1994strong,
  title={Strong evidence for universals in facial expressions: a reply to Russell's mistaken critique.},
  author={Ekman, P.},
  journal={Psychological Bulletin},
  year={1994},
  volume={115 2},
  pages={
          268-87
        }
}

@article{Ekman2002,
  title={Facial action coding system: The manual on CD ROM},
  author={Ekman, P. and Friesen, W. V. and Hager, J. C.},
  journal={A Human Face, Salt Lake City},
  pages={77--254},
  year={2002},
  publisher={A Human Face UT}
}

@book{Orozco2016,
  title={Analysis of speech of people with Parkinson's disease},
  author={Orozco-Arroyave, J.R.},
  year={2016},
  publisher={Logos-Verlag, Berlin}
}

@article{orozco2018neurospeech,
  title={NeuroSpeech: An open-source software for Parkinson's speech analysis},
  author={Orozco-Arroyave, J. R. and Vásquez-Correa, J. C. and Vargas-Bonilla, J. F. and Arora, R. and Dehak, N. and Nidadavolu, P. S. and others},
  journal={Digital Signal Processing},
  volume={77},
  pages={207--221},
  year={2018},
  publisher={Elsevier}
}

@article{Gait,
 author = {Ga{\ss}ner, H. and Steib, S. and Klamroth, S. and Pasluosta, C.F. and Adler, W. and Eskofier, B.M. and Pfeifer, K. and Winkler, J. and Klucken, J.},
 journal = {Journal of Parkinson's Disease},
 pages = {413--426},
 title = {{Perturbation} {Treadmill} {Training} {Improves} {Clinical} {Characteristics} of {Gait} and {Balance} in {Parkinson}'s {Disease}},
  volume={9},
  number={2},
 year = {2019}
}

@ARTICLE{hands,
AUTHOR={Spasojevi\'c, S. and Ili\'c, T. V. and Stojkovi\'c, I. and Potkonjak, V. and Rodić, A. and Santos-Victor, J.},   
TITLE={Quantitative Assessment of the Arm/Hand Movements in Parkinson’s Disease Using a Wireless Armband Device},      
JOURNAL={Frontiers in Neurology},      
VOLUME={8},
NUMBER={388},
PAGES={1--15},     
YEAR={2017},      
}

@article{handwriting,
 author = {Rios-Urrego, C. D. and Vásquez-Correa, J. C. and Vargas-Bonilla, J. F. and Nöth, E. and Lopera, F. and Orozco-Arroyave, J. R. },
 journal = {Computer Methods and Programs in Biomedicine},
 pages = {43--52},
 title = {Analysis and evaluation of handwriting in patients with Parkinson's disease using kinematic, geometrical, and non-linear features},
 volume = {173},
 year = {2019}
}

@inproceedings{grammatikopoulou2019detecting,
  title={Detecting hypomimia symptoms by selfie photo analysis: for early Parkinson disease detection},
  author={Grammatikopoulou, A. and Grammalidis, N. and Bostantjopoulou, S. and Katsarou, Z.},
  booktitle={Proceedings of the 12th ACM International Conference on PErvasive Technologies Related to Assistive Environments},
  pages={517--522},
  year={2019}
}

@article{Vasquez2019,
  title={Multimodal assessment of Parkinson's disease: a deep learning approach},
  author={Vásquez-Correa, J.C. and Arias-Vergara, T. and Orozco-Arroyave, J.R. and Eskofier, B. and Klucken, J. and N\"oth, E.},
  journal={IEEE Journal of Biomedical and Health Informatics},
  volume={23},
  number={4},
  pages={1618--1630},
  year={2019}
}

@article{simons2004emotional,
  title={Emotional and nonemotional facial expressions in people with Parkinson's disease},
  author={Simons, G. and Pasqualini, M. C. S. and Reddy, V. and Wood, J.},
  journal={Journal of the International Neuropsychological Society},
  volume={10},
  number={4},
  pages={521--535},
  year={2004},
  publisher={Cambridge University Press}
}

@article{parnandi2017visual,
  title={Visual biofeedback and game adaptation in relaxation skill transfer},
  author={Parnandi, A. and Gutierrez-Osuna, R.},
  journal={IEEE Transactions on Affective Computing},
  volume={10},
  number={2},
  pages={276--289},
  year={2017},
  publisher={IEEE}
}

@ARTICLE{Li2019vision,  
author={Li, K. and Tao, W. and Liu, L.},  
journal={IEEE Access},  
title={Online Semantic Object Segmentation for Vision Robot Collected Video},   
year={2019},  
volume={7},  
pages={107602-107615},
}

@article{celiktutan2017multimodal,
  title={Multimodal human-human-robot interactions (mhhri) dataset for studying personality and engagement},
  author={Celiktutan, O. and Skordos, E. and Gunes, H.},
  journal={IEEE Transactions on Affective Computing},
  year={2017},
  publisher={IEEE}
}

@article{pampouchidou2017automatic,
  title={Automatic assessment of depression based on visual cues: A systematic review},
  author={Pampouchidou, A. and Simos, P. and Marias, K. and Meriaudeau, F. and Yang, F. and Pediaditis, M. and Tsiknakis, M.},
  journal={IEEE Transactions on Affective Computing},
  year={2017},
  publisher={IEEE}
}

@article{hamm2011automated,
  title={Automated facial action coding system for dynamic analysis of facial expressions in neuropsychiatric disorders},
  author={Hamm, J. and Kohler, C. G. and Gur, R. C. and Verma, R. },
  journal={Journal of Neuroscience Methods},
  volume={200},
  number={2},
  pages={237--256},
  year={2011},
  publisher={Elsevier}
}

@article{wu2014objectifying,
  title={Objectifying facial expressivity assessment of Parkinson’s patients: preliminary study},
  author={Wu, P. and Gonzalez, I. and Patsis, G. and Jiang, D. and Sahli, H. and Kerckhofs, E. and Vandekerckhove, M.},
  journal={Computational and Mathematical Methods in Medicine},
  year={2014},
  publisher={Hindawi}
}

@article{kang2019voluntary,
  title={Voluntary and spontaneous facial mimicry toward other’s emotional expression in patients with Parkinson’s disease},
  author={Kang, J. and Derva, D. and Kwon, D. Y. and Wallraven, C.},
  journal={PloS One},
  volume={14},
  number={4},
  year={2019},
  publisher={Public Library of Science}
}

@incollection{lim2013prevalence,
  title={Prevalence and correlates of complete mental health in the South Korean adult population},
  author={Lim, Y. J. and Ko, Y. G. and Shin, H. C. and Cho, Y},
  booktitle={Mental Well-Being},
  pages={91--109},
  year={2013},
  publisher={Springer}
}

@article{Argaud2018,
  title={Facial Emotion Recognition in Parkinson's Disease:
A Review and New Hypotheses},
  author={Argaud, S. and V\'erin, M. and Sauleau, P. and Grandjean, D.},
  journal={Movement Disorders},
  volume={33},
  number={4},
  pages={554--567},
  year={2018},
  publisher={Wiley Online Library}
}

@article{Sonawane2020,
    title={Review of automated emotion-based quantification of facial expression in Parkinson's patients},
    author={Sonawane, B. and Sharma, P.},
    journal={The Visual Computer},
    year={2020},
    month={Jun},
    day={08},
}

@article{moro2019phonetic,
  title={Phonetic relevance and phonemic grouping of speech in the automatic detection of Parkinson’s Disease},
  author={Moro-Velazquez, L. and Gomez-Garcia, J. A and Godino-Llorente, J. I. and Grandas-Perez, F. and Shattuck-Hufnagel, S. and Yag{\"u}e-Jimenez, V. and Dehak, N.},
  journal={Scientific Reports},
  volume={9},
  number={1},
  pages={1--16},
  year={2019},
  publisher={Nature Publishing Group}
}

@article{dentamaro2020gait,
  title={Gait Analysis for Early Neurodegenerative Diseases Classification Through the Kinematic Theory of Rapid Human Movements},
  author={Dentamaro, V. and Impedovo, D. and Pirlo, G.},
  journal={IEEE Access},
  volume={8},
  pages={193966--193980},
  year={2020},
  publisher={IEEE}
}

@article{de2019handwriting,
  title={Handwriting analysis to support neurodegenerative diseases diagnosis: A review},
  author={De Stefano, C. and Fontanella, F. and Impedovo, D. and Pirlo, G. and di Freca, A. S.},
  journal={Pattern Recognition Letters},
  volume={121},
  pages={37--45},
  year={2019},
  publisher={Elsevier}
}

@book{singh2020domain,
  title={Domain Adaptation for Visual Understanding},
  author={Singh, R. and Vatsa, M. and Patel, V. M and Ratha, N.},
  year={2020},
  publisher={Springer}
}

\vspace{10mm}

\begin{wrapfigure}{l}{25mm} 
\includegraphics[width=1in,height=1.25in,clip,keepaspectratio]{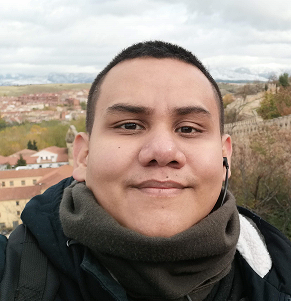}
\end{wrapfigure}\par
\textbf{Luis Felipe Gómez-Gómez} received the B.S. degree in Telecommunications Engineering 
from Universidad de Antioquia, Medellin, Colombia in 2018. Currently, 
he is a Master student at the GITA Lab from the Universidad de Antioquia. 
He has performed research activities related to signal processing and machine learning for biometric applications during the last three years, both with academic and industrial partners. 
His research interests include image processing, signal processing, pattern recognition, 
machine learning, deep learning, biometrics signal processing and their applications in health-care.\par

\vspace{5mm}

\begin{wrapfigure}{l}{25mm} 
\includegraphics[width=1in,height=1.25in,clip,keepaspectratio]{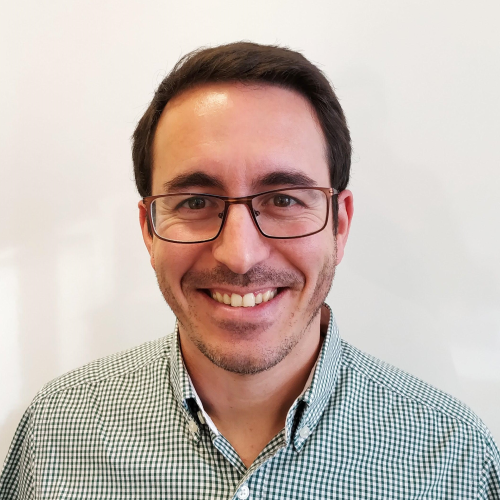}
\end{wrapfigure}
\textbf{Aythami Morales} received the M.Sc. degree in
Telecommunication Engineering from the Universidad
de Las Palmas de Gran Canaria, Spain in 2006 and
the Ph.D. degree from La Universidad de Las Palmas
de Gran Canaria in 2011. Since 2017, he is
Associate Professor affiliated to the Biometrics and Data Pattern Analytics - BiDA Lab at the Universidad Autónoma de Madrid. 
His research interests are focused on pattern recognition, computer vision,
machine learning, and biometrics signal processing. \par

\vspace{5mm}

\begin{wrapfigure}{l}{25mm} 
\includegraphics[width=1in,height=1.25in,clip,keepaspectratio]{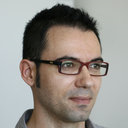}
\end{wrapfigure}\par
\textbf{Julián Fierrez} received the M.Sc. and Ph.D. degrees
in telecommunications engineering from the
Universidad Politécnica de Madrid, Spain, in 2001
and 2006, respectively. Since 2004 he is
affiliated to the Biometrics and Data Pattern Analytics - BiDA Lab at the Universidad Autónoma de Madrid where he is Associate Professor . 
His research interests include signal and
image processing, pattern recognition, and biometrics, with an emphasis on
multibiometrics, biometric evaluation, system security, forensics, and mobile
applications of biometrics.\par

\vspace{5mm}
\vfill\null

\begin{wrapfigure}{l}{25mm} 
\includegraphics[width=1in,height=1.25in,clip,keepaspectratio]{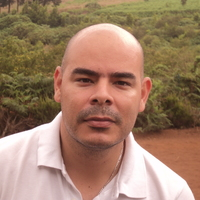}
\end{wrapfigure}\par
\textbf{Juan Rafael Orozco-Arroyave} received the B.S. degree in Electronics Engineering and the
M.Sc. degree in Telecommunications Engineering from Universidad de Antioquia, Medellin,
Colombia, in 2004 and 2011, respectively. He received the Ph.D. degree in Computer Science from the Friedrich-Alexander-Universit\"{a}t Erlangen-N\"{u}rmberg, Germany, in 2015. He is currently an Associate Professor and the head of the GITA Lab at the Universidad de Antioquia, and adjunct researcher with the Pattern Recognition Lab at the 
Friedrich-Alexander-Universit\"{a}t Erlangen-N\"{u}rmberg. His main research 
interests include speech and language processing, signal processing, pattern recognition, 
machine learning, and their applications to different fields in medicine.\par

\end{document}